\documentclass{article}

\usepackage[nonatbib, preprint]{neurips_2021}

\usepackage[utf8]{inputenc} 
\usepackage[T1]{fontenc}    
\usepackage{hyperref}       
\usepackage{url}            
\usepackage{booktabs}       
\usepackage{amsfonts}       
\usepackage{nicefrac}       
\usepackage{microtype}      
\usepackage{xcolor}         

\usepackage{subfigure} 
\usepackage{placeins}
\usepackage{algorithm}
\usepackage{algorithmic}
\usepackage{array}
\usepackage{overpic}
\usepackage{wrapfig}
\usepackage{bm}
\usepackage{comment}
\usepackage{pgfplots}
\usepackage{pgf}
\usepackage{tikz}
\usepackage{soul}
\usepackage{tabstackengine}
\usepackage{mathtools}

\renewcommand{\S}{\mathcal{S}}
\newcommand{\T}{\mathcal{T}}

\newcommand{\M}{\mathcal{M}}

\definecolor{mygray}{RGB}{120,120,120} 

\title{Shape registration in the time of transformers}

\author{%
  Giovanni Trappolini \\
  Department of Computer Engineering\\
  Sapienza University of Rome\\
  \And
  Luca Cosmo\\
  Department of Computer Science\\
  Sapienza University of Rome\\
  \And
  Luca Moschella\\
  Department of Computer Science\\
  Sapienza University of Rome\\
  \And
  Riccardo Marin\\
  Department of Computer Science\\
  Sapienza University of Rome\\
  \And
  Simone Melzi\\
  Department of Computer Science\\
  Sapienza University of Rome\\
  \And
  Emanuele Rodolà\\
  Department of Computer Science\\
  Sapienza University of Rome\\
}

\begin{document}
\maketitle

\begin{abstract} 
In this paper, we propose a transformer-based procedure for the efficient registration of non-rigid 3D point clouds. The proposed approach is data-driven and adopts for the first time the transformer architecture in the registration task. 
Our method is general and applies to different settings. Given a fixed template with some desired properties (e.g. skinning weights or other animation cues), we can register raw acquired data to it, thereby transferring all the template properties to the input geometry. 
Alternatively, given a pair of shapes, our method can register the first onto the second (or vice-versa), obtaining a high-quality dense correspondence between the two.
In both contexts, the quality of our results enables us to target real applications such as texture transfer and shape interpolation.
Furthermore, we also show that including an estimation of the underlying density of the surface eases the learning process. By exploiting the potential of this architecture, we can train our model requiring only a sparse set of ground truth correspondences ($10\sim20\%$ of the total points). 
The proposed model and the analysis that we perform pave the way for future exploration of transformer-based architectures for registration and matching applications. Qualitative and quantitative evaluations demonstrate that our pipeline outperforms state-of-the-art methods for deformable and unordered 3D data registration on different datasets and scenarios.
\end{abstract}


\section{Introduction}
\label{sec:introduction}
\begin{figure}[t!]
\begin{center}
\begin{overpic}
[trim=0cm 0cm 0cm 0cm,clip,width=1.0\linewidth,height=6cm]{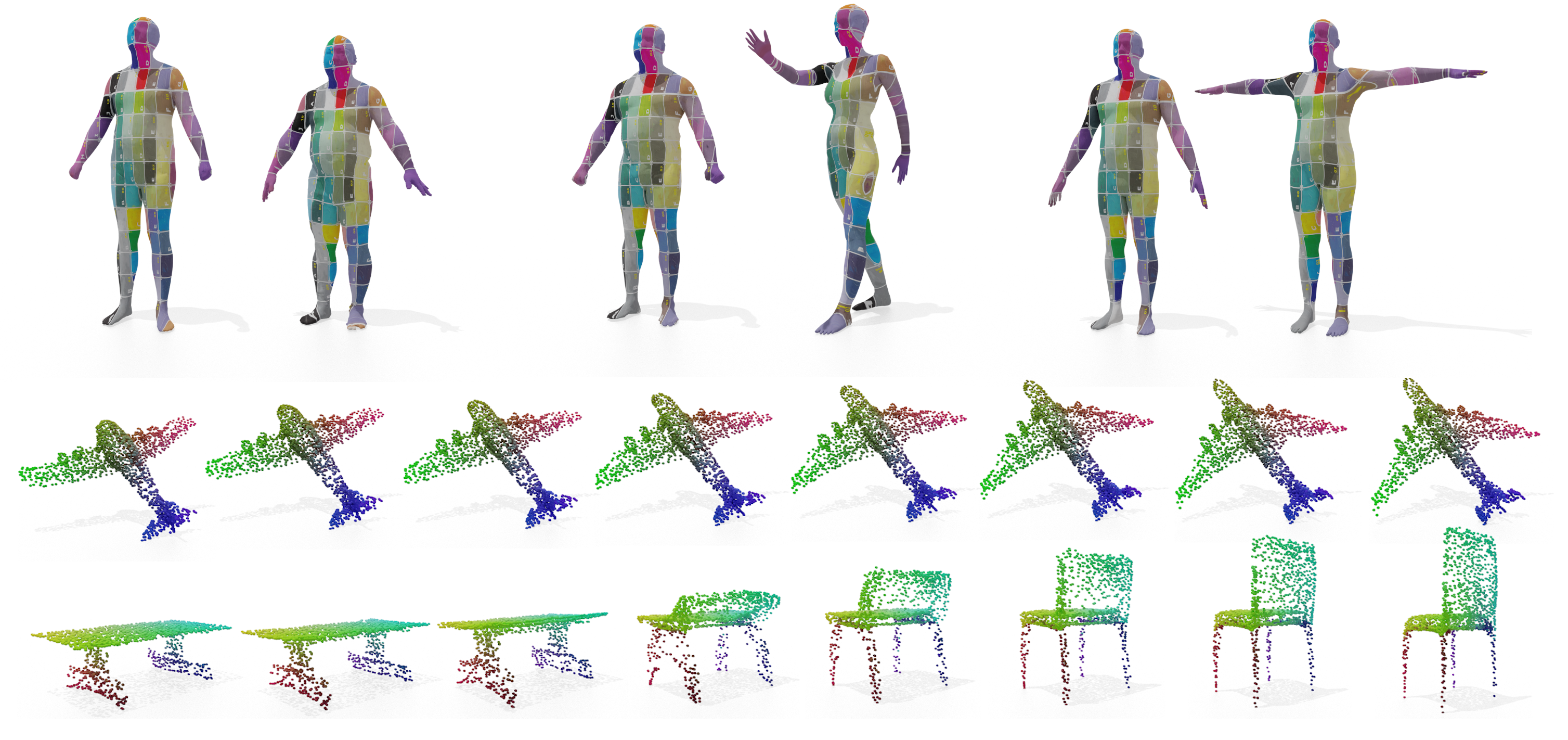}
\put(5.3,75){\footnotesize \emph{source} $\S$}
\put(36,75){\footnotesize \emph{target} $\T$}
\put(69,75){\footnotesize output}
\end{overpic}
\end{center}
\caption{\label{fig:teaser} 
%
Some results in real application targeted by our method. In the first row, three examples of texture transfer on human shape pairs. In the second and third row, two examples of interpolation between intra-class and inter-class shapes from ShapeNet (one for each row).
}
\end{figure}

Recent technological advancements of $3$D acquisition pipelines have produced an abundance of available data. The direct consequence is the non-standardization of the acquisition process. Such technological democratization brings along a disparate amount of different representations, discretization, and arbitrary resolution. Given so, the request to align such data has become urgent. Furthermore, data-driven statistical approaches require aligned data to relate feature changes across the population, inferring underlying patterns.

The Computer Vision community has devoted an extraordinary effort in the last decades to address $3$D objects analysis. A common way to approach this problem is to align the geometry of one known shape to an incoming one. Such methodology is referred to as \emph{registration}. Many different axiomatic pipelines have been proposed that address different kind of objects and domains. While many methods rely on the assumption that the shapes used in the registration task differ just by a rigid transformation, the non-rigid domain is far more complex and interesting. This category pertains organic objects (e.g., humans, animals, internal organs), which are particularly interesting as well. Non-rigid registration aims to align two geometries that may differ by bending and stretching of the geometry, which may also significantly modify its metric. This problem is even more complicated if the geometry representation is given just by a sparse point cloud.

However, an emerging field merges data with classical algorithmic problems, exploiting such statistics as regularization. Among the different learning approaches, recently, the use of the Attention mechanism has become significantly popular in NLP domain, being later transferred to Computer Vision applications. Such architecture is called Transformers, and they represent one of the most significant groundbreaking methodological advancement since the introduction of CNNs.

In this work, we aim to let non-rigid registration meet the transformers. Intuitively, we aim to use the transformer as a \emph{geometrical translator} between two non-rigid point clouds. As the first element, we modified the attention mechanism, proposing to make it aware of the underlying density of the geometry. Hence, we apply such a mechanism in an autoencoder--like architecture, which takes a template point cloud as input and aims to modify its geometry to fit the target point cloud. 

The proposed method achieves better results than several state-of-the-art competitors in the shape matching task. We show results on humans case, but also inter-class objects. In this second case, our method is trained in an unsupervised manner, showing the power of our attention mechanism to infer the underlying geometry. Also, thanks to the attention mechanism, we are able to interpret what the network considers relevant for the registration. Finally, we can target texture transfer and shape interpolation showing applicability in real tasks as demonstrated in \ref{fig:teaser}.
Our contributions could be summarized as follows:
\textbf{(a)} We propose the first transformer for non-rigid registration task, showing the advantages of translation paradigm;
\textbf{(b)} We modified the attention mechanism to make it aware of the point cloud density and of the underlying geometry of a shape;
\textbf{(c)} We significantly improve the state-of-the-art performances on different datasets and challenging scenarios.



\section{Related work}
\label{sec:related}
Shape matching is a problem with a tradition of decades. For a complete overview, we refer to the surveys \cite{van2011survey,sahilliouglu2020recent}; below we cover the literature that more closely relates to our work.

\vspace{1ex}\noindent\textbf{Surface matching}
Early attempts to match non-rigid objects work under the assumption of near isometry. Such is the case, for instance, of blended intrinsic maps \cite{kim2011blended}, which combine multiple conformal mappings with an additional penalty to preserve local areas. Similarly, several variants and applications of the functional maps framework \cite{ovsjanikov2012functional} implicitly assume near isometries by requiring a special structure of the functional representation, or by means of dedicated regularizers \cite{ovsjanikov2017computing,nogneng2017informative,ezuz2017deblurring,melzi2019zoomout,MapTree}. A common drawback to these works is that they do not disambiguate the intrinsic shape symmetries; further, the surface connectivity introduces a structural bias which may affect the performance, as recently shown in \cite{SHREC19}. An attempt to overcome these issues was proposed in \cite{CMH}, but with the extra assumption that the shapes to match are in the same pose.
More recently, SmoothShells \cite{eisenberger2020smooth} proposes an iterative algorithm to recover dense correspondences by an alignment of intrinsic information. While these methods do not address $3$D shape registration directly, they rely on the general idea that a correspondence can be recovered by aligning specialized, possibly high-dimensional embeddings of the shapes at hand. 

\vspace{1ex}\noindent\textbf{Template-free registration}
A popular approach to solve for a matching between $3$D objects is to align their geometries extrinsically via ICP-like procedures \cite{ICP,NRICP,amberg2007optimal}. In fact, registration and matching are intimately related problems with different goals. While the matching problem aims to find a \emph{combinatorial} solution, which indicates for each point its image on the target shape, registration looks for a \emph{spatial} transformation of the geometry. If the two shapes have a significantly different discretization, the latter problem is less ambiguous than the former. ICP-based approaches iteratively solve the two subproblems in an alternating fashion, by finding a point-to-point correspondence and the best transformation that adheres to such correspondence. These methods do not convergence to a good solution if the input shapes are significantly misaligned. Similarly, Coherent Point Drift \cite{myronenko2010point} and variants \cite{jian2010robust,hirose2020CPDbayesian} rephrase the registration problem as an alignment of probability densities. 

\vspace{1ex}\noindent\textbf{Template-based registration}
A different family of approaches make use of a given template, which is known a priori and is possibly parametric, toward which a given input shape is to be matched. 
Taking as an example the case of human bodies, it is common to model the surface deformation related to the subject identity using PCA \cite{allen2003space,anguelov2005scape,loper2015smpl,pavlakos2019expressive}, while recent advancements in statistical data-science suggested that non-linear methods are more expressive to catch fine details of humans \cite{COMA:ECCV18,xu2020ghum,cheng2019meshgan}. Similarly, the pose can be modeled by simple paradigms like Linear-Blend Skinning \cite{loper2015smpl,pavlakos2019expressive}, triangle deformations \cite{anguelov2005scape,hirshberg2012coregistration}, but also learning methods \cite{xu2020ghum}. Efforts to register such templates to arbitrary target models have been carried out extensively by the community  \cite{hirshberg2012coregistration,zuffi2015stitched,FARM,marin2019high}. 
However, the requirement of a template is not always easy to satisfy.

\vspace{1ex}\noindent\textbf{Learning methods}
With the rise of learning methods, several attempts have been made to introduce a statistical prior to the matching and registration process. For example, several extensions have been proposed to bring the functional maps formalism into a learning paradigm \cite{litany2017deep,donati2020deep,sharp2020diffusion}. SmoothShells has also been extended to be data-driven \cite{eisenberger2020deep}. The point cloud representation has received comparably less attention, mainly in a rigid alignment setting  \cite{Huang_2020_CVPR,pais20203dregnet,sarode2019pcrnet}. In the non-rigid domain, a seminal work is 3DCoded \cite{groueix20183d}, that proposes a proper registration using an autoencoder architecture. However, having a fixed template forbids inter--class operations and limits the use of the point cloud structure. Recently, it has been proposed to learn a linearly-invariant embedding \cite{LIE2020}, but the method requires training two separate networks and relies on simple PointNets \cite{qi2017pointnet} which are not able to catch the fine details of the objects. Finally, a recent trend in geometric deep learning suggests that implicit representations may also be used for shape matching \cite{bhatnagar2020loopreg}.


\vspace{1ex}\noindent\textbf{Transformer-based architectures}
Transformers have been first introduced in the context of neural machine translation by the pioneering work  \cite{AttentionIsAllYouNeed}, and later procedeed to revolutionize the field of natural language processing \cite{devlin-etal-2019-bert,radford2018improving}.
The success obtained in NLP inspired further work to employ transformers in computer vision \cite{dosovitskiy2020image}, where they managed to outperform convolutional networks.
Exploiting the input invariance property characterising the attention mechanism, many works have naturally extended transformers to handle point clouds \cite{pct_guo,zhao2020point,engel2020point}. These works show promising results, but work only in the context of object classification and segmentation. Instead, \cite{wang2019deep} proposes a network to find the rigid alignment between two point clouds imitating ICP, and using a transformer architecture to infer the residual term. Differently, we aim to solve for \emph{non}-rigid registration, which is a more general case.





\section{Notation and general objective}
\label{sec:notation}

\noindent\textbf{3D shapes}
We model 3D shapes as compact $2$-dimensional Riemannian manifolds $\M$, possibly with boundary $\partial\M$. In the discrete setting, we represent the manifold $\M$ as an unorganized point cloud of $n_{\M}$ vertices embedded in $\mathbb{R}^3$, and encoded in a coordinate matrix $X_{\M} \in \mathbb{R}^{n_{\M}\times 3}$.

\setlength{\columnsep}{5pt}
\setlength{\intextsep}{1pt}
\begin{wrapfigure}[7]{r}{0.35\linewidth}
\vspace{-0.5cm}
\begin{center}
\begin{overpic}
[trim=0cm 0cm 0cm 0cm,clip,width=1.0\linewidth]{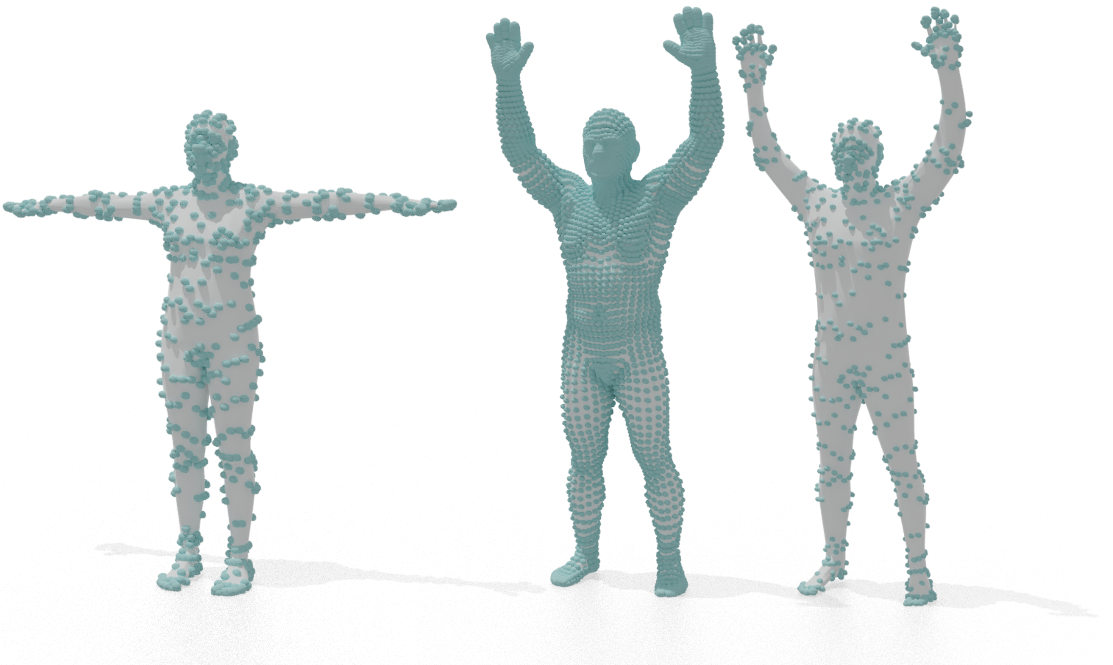}
\put(10,0){\footnotesize {source}}
\put(47,0){\footnotesize {target}}
\put(67,0){\footnotesize registration}
\end{overpic}
\end{center}
\end{wrapfigure}
%

\vspace{1ex}\noindent\textbf{Shape registration}
The main objective of this paper is to introduce a data-driven approach to perform shape registration. Given a \emph{source} shape $\S$ and a \emph{target} shape $\T$, respectively represented by the sets of vertices $X_{\S}$ and $X_{\T}$, our goal is to find a corresponding 3D position for each point in $\S$ on the surface of $\T$. 
An example is shown in the inset figure, where the underlying surface is visualized for reference only.

Our method does {\em not} assume the source $\S$ to be a fixed template, but can generalize to arbitrary shape pairs $\S$ and $\T$; this is in contrast, e.g., with~\cite{groueix20183d}, where the objective is to learn how to deform a fixed known template into another shape.

\vspace{1ex}\noindent\textbf{Attention}
One of the most influential ideas in the recent years, attention originated in the realm of natural language processing but has since gained traction in other fields, such as computer vision and signal processing, due to the vast increase in performance and interpretability exhibited in several tasks.
At its heart, the attention mechanism allows learning models to encode latent relations between inputs, assigning higher importance, or "attention", to the parts deemed more relevant. In this sense, attention allows to efficiently capture context information as well as higher order dependencies.

Formally, given two generic input sequences $X_{1} \in \mathbb{R}^{n \times d}$ and $X_{2} \in \mathbb{R}^{m \times d}$ (for clarity of exposition we assume a constant embedding dimension $d$, but it is not a necessary assumption), the attention mechanism models the linearly encoded representation of the inputs as triplets of query, key, and value matrices, respectively $Q \in \mathbb{R}^{n \times d}$, $K \in \mathbb{R}^{m \times d}$, and $V \in \mathbb{R}^{m \times d}$.
The attention score $W$ is then defined as 
$W = \mathrm{softmax}\left( Q K^{T} d^{-0.5} \right)$ 
%
and used to compute the weighted mean of the value vectors, resulting in an output feature matrix $A=WV$.
%
%
%
%
%
We refer to {\em self}--attention 
whenever $X_{1}$ and $X_{2}$ are the same object, and use the term {\em cross}--attention otherwise.



\section{Method}
\label{sec:method}

Our method takes as input two point clouds $X_{\T}$ and $X_{\S}$, with $n_{\T}$ and $n_{\S}$ points respectively, and deforms the points of the source shape $X_{\S}$ to fit the geometry described by the target point cloud $X_{\T}$. To do so, we rely on a novel attention mechanism which considers the underlying geometry conveyed by the point cloud, rather than treating the points simply as elements of a set.


\vspace{1ex}\noindent\textbf{Surface Attention}
The classic attention definition, as introduced in Section~\ref{sec:notation}, looks at the input data points simply as elements of a set, and uses the computed attention scores to perform a weighted sum of the value vectors.
When the value vectors represent a sampling of a signal over a surface, however, the natural domain for the integration should be the surface itself, of which the weighted sum is just an approximation highly sensitive to the specific surface sampling (depending, for instance, on the acquisition method).

To overcome this limitation we propose to modify the attention mechanism to consider the portion of surface represented by each point, weighting the attention score by an estimated local area element.
In practice, for each point $x_i \in X$, we estimate its area contribution as the inverse of the local point density: $\mathcal{A}(X)_i = (\vert \{x_j \in X ~\mathrm{s.t.}~ \|x_j-x_i\|_2 < r\} \vert)^{-1}$,
where $\vert \cdot \vert$ denotes the cardinality of a set, and $r$ is a local radius ($r=0.05$ in our experiments). Note that, we are not interested in the absolute value of the area elements, rather on the relative contribution of each point. 
The surface attention score is thus defined as:
\begin{equation}
    \widetilde{W}_{i,j} = \frac{e^{W_{i,j}} \mathcal{A}(X)_j}{\sum_t e^{W_{i,t}} \mathcal{A}(X)_t} \,.
\end{equation}
As in the classical attention, output features are computed as $\widetilde{W}V$.
Figure \ref{fig:surface_attn} shows that the surface attention mechanism results in a more stable localization of the attention scores across different samplings of the surface.
In our architecture, each time attention is computed over point clouds, we use such formulation.


\begin{figure}[t!]
\begin{center}
\begin{overpic}
[trim=0cm 0cm 0cm 0cm,clip, width=1.0\linewidth]{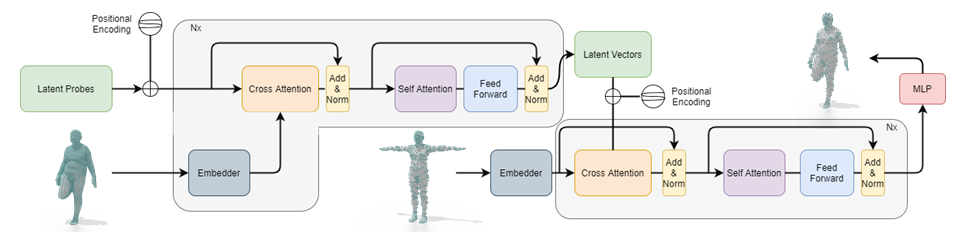}   
\put(6.5,-2){\footnotesize $X_{\T}$}
\put(42.5,-2){\footnotesize $X_{\S}$}
\put(68.5,17.7){\footnotesize $LS_{\T}$}
\end{overpic}
\end{center}
\caption{\label{fig:model} The proposed transformer-based architecture for 3D point cloud registration. The latent probes capture the geometry of the input target shape $X_{\T}$ through the encoder layers. The resulting latent vectors drive the deformation of an input source shape $X_{\S}$ in the decoding layers, resulting in a deformation of the points of $X_{\S}$ to fit the geometry of $X_{\T}$.
}
\end{figure}

\vspace{1ex}\noindent\textbf{Architecture}
The architecture we propose, portrayed in Figure~\ref{fig:model}, is an \emph{iteratively conditioned} autoencoder. The two main ingredients are an encoder, that maps a target point cloud $X_{\T}$ into a latent space $LS_{\T}$, and a decoder, that deforms a source point cloud $X_{\S}$ to resemble the geometry of the target point cloud $X_{\T}$. 

The encoder draws inspiration from  \cite{jaegle2021perceiver}, featuring a set of learnable latent probes $LP$.
It presents multiple layers of cross attention which iteratively condition the latent probes $LP$ with the embedding of $X_{\T}$. 
After each conditioning, the resulting latent space is further transformed by layers of self attention, feed forwards and residual connections. The encoder output is a set of latent vectors $LS_T$ containing relevant information collected from the target point cloud $X_{\T}$.

The decoder is analogous to the encoder but the relationship between the latent space and the point cloud in the cross attention is reversed. 
That is, the embedding of the source point cloud $X_{\S}$ is transformed and iteratively conditioned with the latent space $LS_T$ produced by the encoder.
This procedure induces a deformation of $X_{\S}$ that, after a final MLP layer, aligns to the points of $X_{\T}$.

\vspace{1ex}\noindent\textbf{Training}
Given a training dataset equipped with a ground-truth correspondence, we train our network in a supervised setting. Starting from a set of shapes in correspondence, we minimize the reconstruction error using the standard reconstruction loss:
\begin{equation}%
    L^{sup} (X_{\S}, X_{\T})= \| X_{\T} - D(X_{\S}, E(X_{\T})) \|_2^2\,.
\end{equation}%
Furthermore, in case a ground-truth correspondence is not available, our network can be trained in an unsupervised way using the Chamfer distance, between $X_{\T}$ and $D(X_{\S}, E(X_{\T}))$, as defined in~\cite{groueix20183d}.



\vspace{1ex}\noindent\textbf{Testing}
At test time, our network can register a point cloud to another instantaneously, with a single forward pass. In the following experiments that involve the computation of matching, the correspondence can be obtained just by looking for the Euclidean nearest-neighbor between $X_\T$ and the output $D(X_{\S}, E(X_{\T}))$ in the $3$D space. 

\vspace{1ex}\noindent\textbf{Refinement}
The peculiar structure of our architecture allows us to refine the output during the testing procedure. This is achieved by minimizing the energy function $\textrm{chamfer}(X_{\T},D(X_{\S}, E(X_{\T})))$ with respect to the latent vectors $LS_{\T}$. In practice, we use the latent vectors produced by a single forward pass as initial guess for $LS_{\T}$, and minimize the previous energy function using Adam optimizer.
%
As we show in Section \ref{sec:results}, this step can significantly improve the shape matching results. We remark that this is possible only thanks to the registration formulation of our approach. Shape alignment in $3$D space can be optimized continuously, while a point-to-point assignment is non-differentiable due to the combinatorial nature of the problem.








\section{Experiments \& Results}
\label{sec:results}

We evaluate the effectiveness of our architecture on a number of challenges.
We begin by analyzing the key components of our model, 
motivating our architectural choices through ablation studies. %
Finally, we present our results in the context of matching, registration, and inter--class registration.

\begin{table}[t!]
\centering
\caption{\label{tab:ablation}Ablation study. We report the relative average matching error (lower is better) with respect to the baseline architecture used in all the experiments (for which the error is set to 1). }
\small
\begin{tabular}{l|ccccc|ccc|ccc}
\toprule
Test Set   & \multicolumn{5}{c|}{\# of latent space vectors} & \multicolumn{3}{c|}{\# E/D layers} & \multicolumn{3}{c}{latent space dimension} \\
 & 32 & 16 & 8 & 4 & 2
 & 4 & 8 & 12
 & 32 & \hspace{1em} 64 \hspace{1em} & 128\\ 
\midrule
SURREAL & 1.00 & 1.00 & 0.93 & 0.97 & 0.91
&1.04 & 1.00 & 0.93
&1.36 & 1.00 & 1.45 \\
FAUST   & 1.00 & 1.01 & 1.24 & 1.30 & 1.29
&1.45 & 1.00 & 1.22
&0.82 & 1.00 & 0.94\\
\midrule
\midrule
Average & 1.00 & 1.01 & 1.08 & 1.13 & 1.10
&1.24 & 1.00 & 1.07
&1.09 & 1.00 & 1.19 \\
\bottomrule
\end{tabular}

\end{table}

\begin{figure}[t!]
\begin{center}
\begin{overpic}
[trim=-5cm 0cm 0cm -5cm, clip, width=1\linewidth]{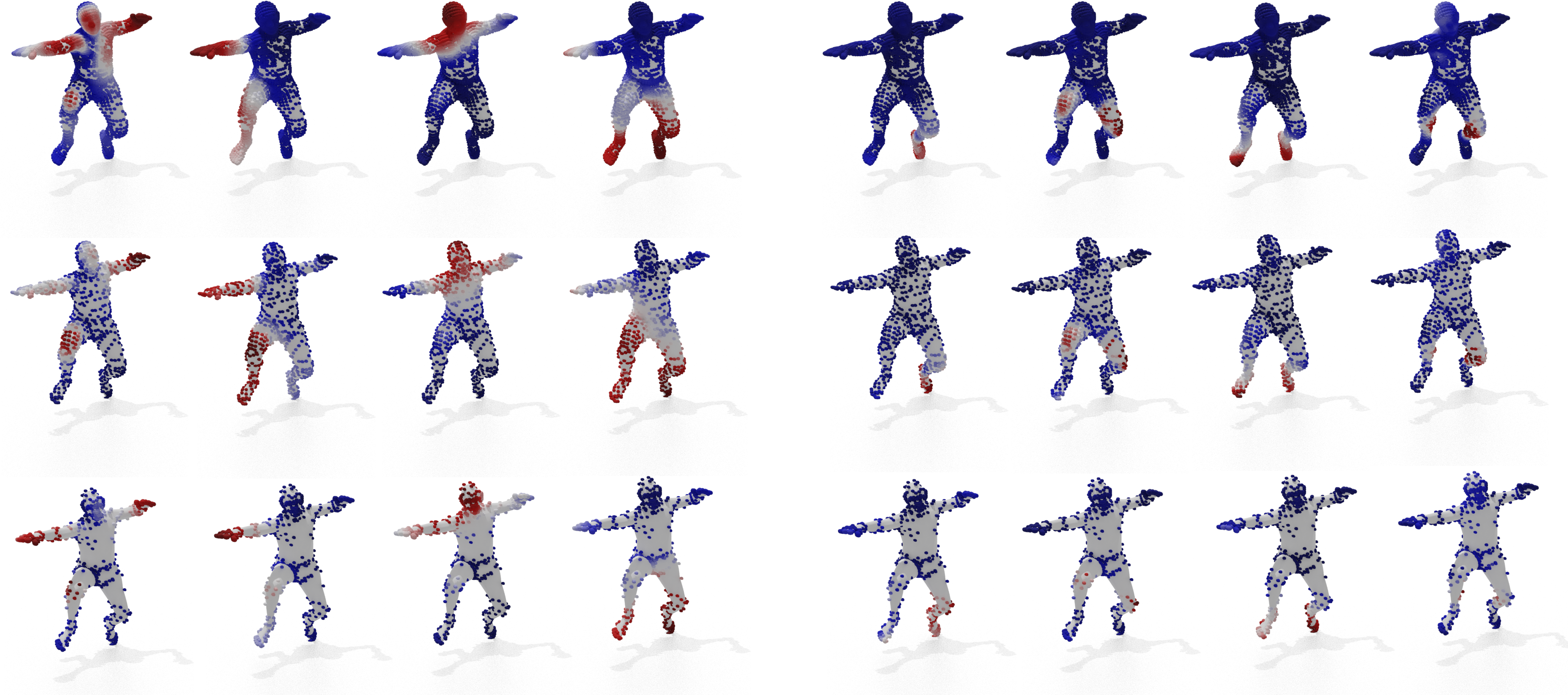}
    \put(21, 44){\footnotesize \emph{Classic attention}}
    \put(71, 44){\footnotesize \emph{Surface attention}}
    \put(0, 36){\scriptsize \rotatebox{90}{{Original}}}
    \put(2, 37.25){\scriptsize \rotatebox{90}{{(7K)}}}
    \put(0, 20){\scriptsize \rotatebox{90}{{Quadric error}}}
    \put(2, 22.5){\scriptsize \rotatebox{90}{{(1K)}}}
    \put(0, 2.5){\scriptsize \rotatebox{90}{{Normal deviation}}}
    \put(2, 7){\scriptsize \rotatebox{90}{{(1K)}}}
\end{overpic}
\end{center}
\caption{\label{fig:surface_attn} Comparison between the classic (left) and surface (right) attention mechanism behavior with differently sampled surfaces. The surface attention is more stable across different sampling strategies. The two attention mechanism have been trained separately, so there is no correspondence on the attention localization between the two. Surfaces are shown just for visualization purposes.}
\end{figure}

\begin{figure}[t!]
\begin{center}
\begin{overpic}
[trim=0cm 0cm 0cm 0cm,clip,width=.8\linewidth]{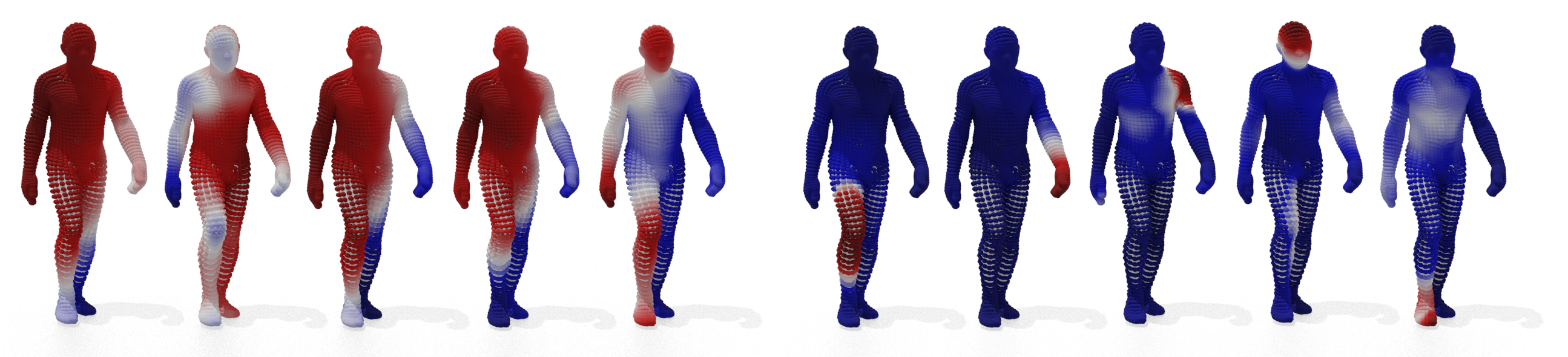}
\end{overpic}
\end{center}
\caption{\label{fig:cross-attn-decoder} The colormap shows the attention value given to each point of the shape by each input latent vector (only 5 of 32 are shown) at the first (left half) and last (right half) layers of the decoder.}
\end{figure}

\subsection{Experimental settings}

\noindent\textbf{Training data}
For all our experiments in the humans domain, we trained our method on the same shapes from the SURREAL dataset \cite{varol17_surreal} used in \cite{LIE2020}. It consists of $10000$ point clouds for training. Each point cloud has 1000 points, which simulate a significantly sparse sampling of the original shape. During training we augment the data by randomly rotating shapes along the second axis. We also trained our model on the ShapeNet dataset \cite{chang2015shapenet}. This dataset is composed of 16881 point clouds representing 3D shapes from 16 different objects categories(from chairs to airplanes), that also do not share a ground truth correspondence. Hence, we trained our method in an unsupervised manner for this particular data. We sample 1024 points for each object and use the same train/test split as in~\cite{wang2019dynamic}.

We also compare our model on other human datasets, thus assessing the ability to generalize to data out of the training distribution.
A popular dataset to analyze real identities and poses is FAUST~\cite{FAUST}, which is composed by ten subjects in ten different poses.
We also used a 1000 points version of it, which we refer to as \emph{FAUST1K}. To simulate the noise produced by a $3$D acquisition pipeline, we considered the same data from \cite{LIE2020} in which the points are perturbated by a Gaussian noise. Also, we challenge our method on \emph{SHREC}'19 \cite{SHREC19}. Such dataset is composed by 44 shapes which have different connectivities, poses and densities. 
In all the experiments we refer to our method as \emph{Our}.

In our comparisons we train our model for 10000 epochs using Adam optimizer \cite{kingmaAdam}. We use 32 latent probes of dimension 64, and 8 layers for both the encoder and the decoder.

\vspace{1ex}\noindent\textbf{Competitors}
We consider 3DCoded  \cite{groueix20183d} (\emph{3DC}) as our principle competitor. Similarly to us, it aims at deforming a shape into another using an autoencoder architecture. However, it assumes one of the two shapes to be a predefined template, limiting its generalization capability. Further, we consider the Linearly-Invariant Embedding approach of \cite{LIE2020} (\emph{LinInv}), which learns an high-dimensional embedding in which the shapes can be aligned by a linear transformation. It is based on PointNet, and do not exploit any local structural mechanism. Finally, we considered the Geometric Functional Maps definition \cite{donati2020deep}, using DiffusionNet \cite{sharp2020diffusion} as feature extractor (\emph{DiffNet}). Similarly to \cite{LIE2020}, it learns to embed each shape point into a common higher-dimensional. 
Both our method and 3DC can improve the registration with a post-processing \emph{refinement}, we refer to these as \emph{3DC$_R$} and \emph{Our$_R$}.

\subsection{Ablation and analysis}
Here we justify our choices on hyperparameters through an ablation study, and we present some insight on the key properties of our method with an in depth analysis.

\vspace{1ex}\noindent\textbf{Ablation study}
To investigate the different components of the proposed architecture, we run a batch of experiments in a reduced version of the SURREAL dataset, using the first 1000 shapes for training, and a different set of 2700 shapes for testing. We also test on FAUST1K.

We report the results in Table~\ref{tab:ablation}. We test for three possible ablations: number of latent space vectors, their dimension, and the number of encoder--decoder blocks.
We remark that the while SURREAL share discretization and density with the training data, FAUST1K does not.
This may explain the difference in performance we observe, and their almost inverse relationship. 
In fact, it seems that a low number of latent vectors tend to overfit the specific sampling seen during training, while a higher number seems to provide better generalization. A similar behavior can be observed for the other hyperparameters considered.
Finally, to reach a good trade off between bias and variance, we choose the configuration that produces the minimum mean error on these two evaluation sets.

\vspace{1ex}\noindent\textbf{Analysis of the Surface attention}
Quantitative and qualitative results showing the importance of the surface attention mechanism compared to the standard point attention. 
This novel kind of attention we propose has the merit of being much more agnostic to the particular discretization and density of the given point cloud.
We can observe this behavior clearly in Figure~\ref{fig:surface_attn}. 
Here we visualize the attention across different densities and discretization strategies and two different settings, one with surface attention and one with regular attention. 
With the regular attention mechanism the part of the point cloud attended shows erratic behavior, with different intensities and often even different part that gets attended.
Surface attention completely solves this issue, enabling the architecture to achieve greater generalization capacity and enjoying increased robustness, and allowing us to decrease the error on the full version of FAUST of more that $50\%$.

\begin{figure}[t!]
\begin{center}
\begin{overpic}
[trim=2cm 0cm 0cm 0cm,clip,width=1\linewidth]{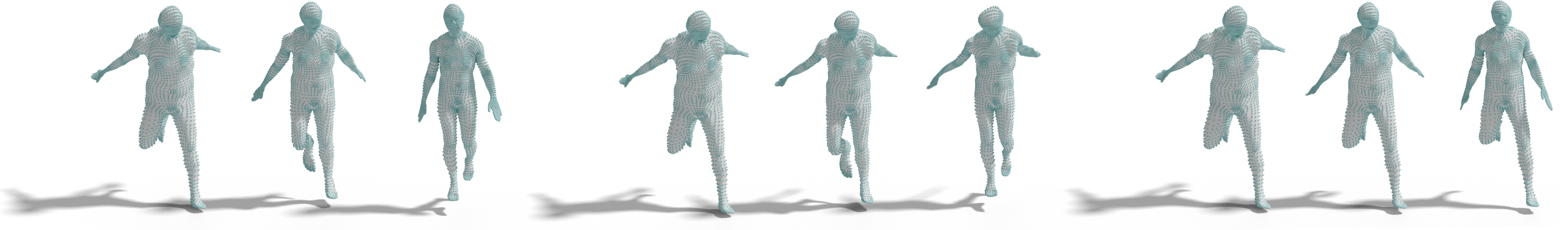}
\end{overpic}
\end{center}
\caption{\label{fig:interpolation-ls} Interpolation examples of two shapes. Left group: left and right-most shapes are the registration of a source shape $S$ to two different target shapes $T_1$, $T_2$, the central shape is obtained passing as input to the decoder latent vectors obtained by linearly interpolating the latent vectors of the two target shapes.
In the second and third groups the interpolation is performed only on a subset of the vectors: we fix the latent vectors of the first shape which attention, on the last layer of the decoder, focuses respectively on the upper part of the body (center) and on the legs (right), while interpolating the remaining ones.}
\end{figure}

\vspace{1ex}\noindent\textbf{Multivariate Latent Space}
Inspecting the cross attention of the decoder layer we can seize the impact of the latent probes in our learning.
In the top row of Figure~\ref{fig:cross-attn-decoder} we can see how the attention in the first layer of the decoder captures global information of the shape, while in the last layer (bottom row), attention puts its focus on small details of the shape.

The possibility of directly visualizing attention maps and the multivariate nature of the latent space permits further analysis.
In Figure~\ref{fig:interpolation-ls}, in particular, we explore the structure of our latent space and how attention influences it.
We grab the two latent spaces output of the encoder by registering a source shape $\S$ to two different target shapes $\T_1, \T_2$. 
Starting from the left triad, the left most and the right most shape represents the registration of $\S$ to $\T_1$ and $\T_2$ respectively; the one in the middle is a obtained by linearly interpolating the two latent spaces.
In the middle triad we add a twist to this procedure, namely we locate the latent vectors attending the most to the upper body of the shape and keep them fixed. A similar procedure is undertaken in the right triad, but this time focusing on the legs.
From this experiments, it follows that our latent space is not only linearly navigable, meaning that a linear interpolation of the shapes encoded in the latent space preserve a reasonable semantics, but also, and most interestingly we might say, attention characterizes this space and directly allow for meaningful alteration, or preservation, of selected chosen characteristics.

\subsection{Results and Applications}
In this section we present results and application of our method. In particular, we show state of the art performance on the shape matching task as well as shape registration.

\begin{figure}[t!]
\begin{center}
\begin{tabular}{lr}
        \hspace{-0.35cm}
        \begin{minipage}{0.23\linewidth}
        \vspace{-3.5cm}
        \input{./figures/tikz/s2s_2.tikz}
        \end{minipage}
        \hspace{0.4cm}
        &
        \begin{overpic}
        [trim=0cm 0cm 0cm 0cm,clip,width=0.73\linewidth]{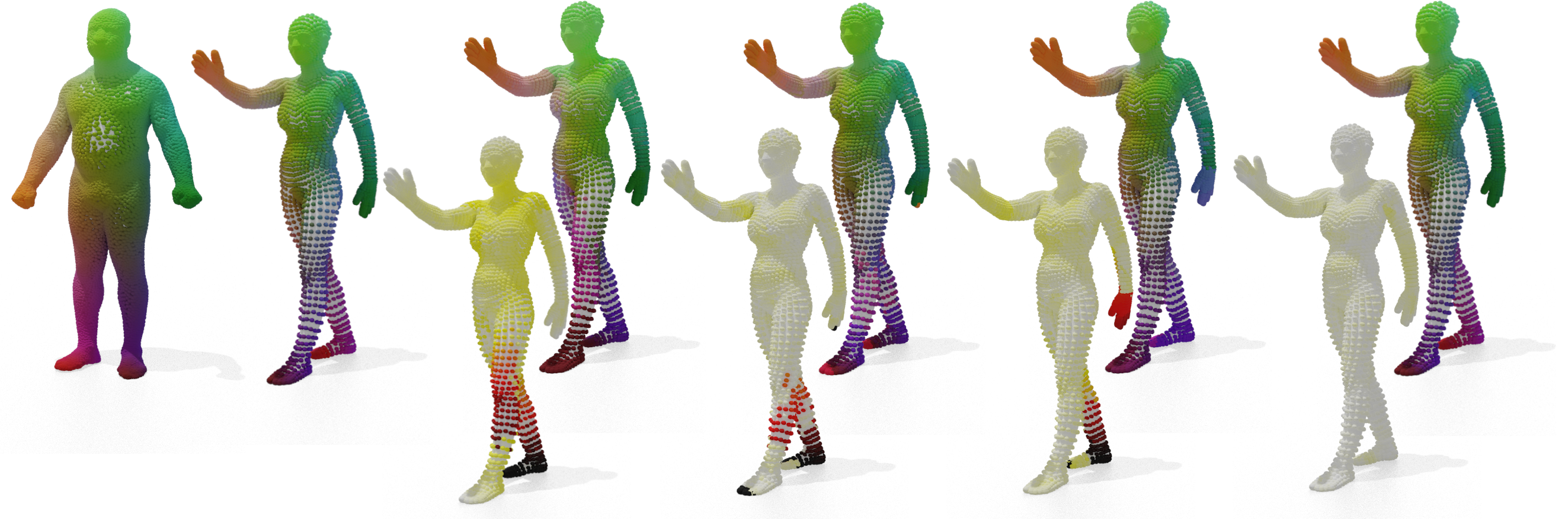} 
        \put(5.1,0.75){\footnotesize Source}
        \put(18,0.75){\footnotesize GT}
        \put(36,0.75){\footnotesize 3DC$_R$}
        \put(54,0.75){\footnotesize DiffNet}
        \put(72,0.75){\footnotesize LinInv}
        \put(90,0.75){\footnotesize Our$_R$}
        \end{overpic}
\end{tabular}
\end{center}
\caption{\label{fig:shrec19} Comparison of different methods on SHREC19 \cite{SHREC19}. \textbf{Left}: Each curve shows the percentage of points (\emph{y-axis}) with at most a geodesic error (\emph{x-axis}). \textbf{Right:}
Qualitative comparison. From left to right, the source shape $\S$, the ground truth color transfer to the target geometry $\T$, the results of the competitors and our result. The color transfer predictions are paired with the corresponding error visualizations, from white (error=0) to black (error>0.75).}
\end{figure}

\begin{figure}[t!]
\begin{center}
\begin{overpic}
[trim=0cm 0cm 0cm 0cm,clip,width=1.0\linewidth]{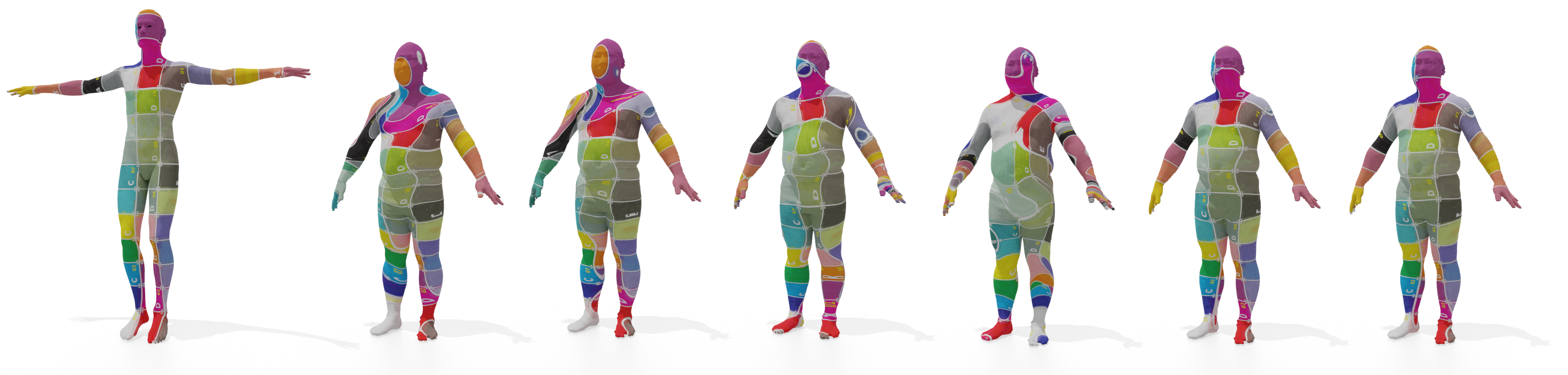}
\put(7,0){\footnotesize Source}
\put(24.2,0){\footnotesize 3DC}
\put(37.5,0){\footnotesize 3DC$_R$}
\put(48,0){\footnotesize DiffNet}
\put(63,0){\footnotesize LinInv}
\put(75,0){\footnotesize Our}
\put(89,0){\footnotesize Our$_R$}
\end{overpic}
\end{center}
\caption{\label{fig:all_texture_transfer} Qualitative comparison of texture transfer on the SHREC19 \cite{SHREC19} dataset. 
From left to right, the source shape $\S$, the texture transfer to the target shape $\T$ of competitors and our results.} 
\end{figure}

\vspace{1ex}\noindent\textbf{Matching}
One of the task we consider is that of matching. Given two generic point clouds we want to find correspondences between them. 
Our model approach this task in a natural and elegant way, by registrating one shape onto the other. It becomes trivial then to obtain correspondences through a nearest point search.
Results are reported in Table \ref{tab:s2s}. Our method consistently outperforms the state of the art by a solid margin. Furthermore, we notice that our method is endowed with a much greater ability to generalize. This can be noted on the SHREC'19 dataset, as visualzied in Figures \ref{fig:shrec19}. The quality of our matching enable us to achieve high quality texture transfer as shown in Figures~\ref{fig:teaser} and \ref{fig:all_texture_transfer}.

\begin{table}[t!]
\begin{center}
\caption{\label{tab:s2s} Comparison of the average geodesic error on different datasets. FAUST(1k) is obtained from FAUST sampling 1k points, FAUST(1k-noise) is obtained as FAUST(1K) but perturbing each vertex with Gaussian noise. FAUST~\cite{FAUST} and SHREC19~\cite{SHREC19} have very different sampling densities, with point clouds ranging from $\sim 5$ to $\sim 200$ thousand points.}
\small
\begin{tabular}{lcccc}
\toprule
Method & FAUST & FAUST(1k) & FAUST(1k-noise) & SHREC19 \\ 
\toprule
3DC               & 0.0776 & 0.0542 & 0.0712 & 0.2138 \\
DiffNet           & 0.0656 & 0.0534 & 0.0985 & 0.1509 \\
LinInv            & 0.0942 & 0.0471 & 0.0618 & 0.1284 \\
\textbf{Our}      & \textbf{0.0513} & \textbf{0.0419} & \textbf{0.0510} & \textbf{0.0802} \\ 
\midrule 
3DC$_R$           & 0.0485 & 0.0367 & 0.0526 & 0.1935 \\
\textbf{Our$_R$}  & \textbf{0.0369} & \textbf{0.0263} & \textbf{0.0410} & \textbf{0.0615} \\
\bottomrule
\end{tabular}
\end{center}
\end{table}
%

%
\vspace{1ex}\noindent\textbf{Template Registration}
A classical problem in Computer Graphics is to register a given template, usually a triangular or polygonal mesh, to some acquired point cloud. This setup is a special instance our method, in which the point cloud to be given as input to the decoder remains constant.
Regarding our competing methods, even if DiffusionNet and LinInv are not proper registration algorithm, we can move a template point to the corresponding point (as found by the matching algorithm) on the input point cloud. This partially explains why, even though performing the worst overall, achieve lower chamfer distance, since their error is somewhat bound.
On the other hand, 3DC is trained exactly in this fashion. Note also that 3DC is the only method that sees the template shape during the training phase.
Even though we train our method on a different task, we manage to improve on the state of the art as can be seen in Table~\ref{tab:s2t}, without the need for any, altough possible, fine-tuning.

\begin{table}[t!]
    \caption{\label{tab:s2t} Comparison on the registration task on FAUST \cite{FAUST}. \textbf{Left}: Each curve shows the percentage of points (\emph{y-axis}) with at most that geodesic error (\emph{x-axis}). \textbf{Right}: Table showing for each method: the mean geodesic error (MGO)  of the resulting matching; the Chamfer distance, the maximum and the mean Euclidean distance (Max EU, Mean EU) between the registered template and the target.}
    \begin{center}
        \hspace{-0.4cm}
        \begin{minipage}{0.23\linewidth}
            \input{./figures/tikz/s2t_2.tikz}
        \end{minipage}
        \hspace{2cm}
        \begin{minipage}{0.60\linewidth}
            {\small
            \begin{tabular}{lcccccccc}
                \toprule
                Method & Chamfer & Max EU & Mean EU & MGO\\
                \toprule
                3DC              & 0.0409            & \textbf{0.2231}   & 0.0723            & 0.0463  \\
                DiffNet          & \textbf{0.0164}   & 1.2942            & 0.1023            & 0.0761  \\
                LinInv           & 0.0177            & 0.3314            & 0.1044            & 0.0692  \\
                \textbf{Our}     & 0.0333            & 0.2299            & \textbf{0.0650}   & \textbf{0.0434}  \\
                \midrule   
                3DC$_R$          & 0.0214            & 0.1705            & 0.0445            & 0.0293 \\
                \textbf{Our$_R$} & \textbf{0.0129}   & \textbf{0.1626}   & \textbf{0.0306}   & \textbf{0.0275} \\
                \bottomrule
            \end{tabular}} 
        \end{minipage}
    \end{center}
\end{table}

\subsection{Unsupervised Registration and Interpolation}
One of our main advantages is that we do not require a template. Fixing a common template is not trivial, if not possible at all, when dealing with very different objects. To show the ability of our method of dealing with this challenging scenario we trained a model to register pair of shapes belonging to possibly different object categories of ShapeNet, using the chamfer loss defined in Section~\ref{sec:method}.
We show in Figure \ref{fig:teaser} (2nd and 3rd row) two interpolation sequences between two airplanes and between a chair and a table, showing that our method is able to register different objects preserving a meaningful correspondence, represented by similar colors.
The interpolated reconstructions are obtained by embedding the outermost shapes in the latent space through the encoder and then using the linearly interpolated latent vectors and the left-most shape as input to the decoder.






\section{Conclusion}
\label{sec:conclusion}

We propose the first transformer based architecture to tackle the problem of non--rigid registration.
We introduce a novel surface attention mechanism better suited to exploit the local geometric priors of the underlying structure. 
Our method reaches state of the art performance in shape matching and shape registration without assuming any fixed template, and  
generalizes also to different and complex geometries, e.g. handling multiple classes of ShapeNet \cite{chang2015shapenet} simultaneously.
The attention mechanism at the core of our architecture has the potential to enforce \emph{local control} of the interpolation, as seen in Figure~\ref{fig:interpolation-ls}.
Further investigation is needed to explore the possibility to introduce additional priors on the attention to force a semantically meaningful localization and interpolation behavior. 
Our method shares a common drawback with most transformed-based architectures, requiring long training and post-processing time due to the nature of the refinement procedure.
All code and data is publicly available \footnote{https://github.com/GiovanniTRA/transmatching}.

\section{Acknowledgment}

This work is supported by the ERC Grant No. 802554 (SPECGEO) and the MIUR under grant “Dipartimenti di eccellenza 2018-2022”.






\bibliographystyle{splncs}
\bibliography{biblio}

\begin{thebibliography}{10}
\providecommand{\url}[1]{\texttt{#1}}
\providecommand{\urlprefix}{URL }
\providecommand{\doi}[1]{https://doi.org/#1}

\bibitem{scantheworld}
Scan the world project. \url{https://www.myminifactory.com/scantheworld},
  [Online; accessed 04-June-2021]

\bibitem{allen2003space}
Allen, B., Curless, B., Popovi{\'c}, Z.: The space of human body shapes:
  reconstruction and parameterization from range scans. ACM transactions on
  graphics (TOG)  \textbf{22}(3),  587--594 (2003)

\bibitem{amberg2007optimal}
Amberg, B., Romdhani, S., Vetter, T.: Optimal step nonrigid icp algorithms for
  surface registration. In: 2007 IEEE conference on computer vision and pattern
  recognition. pp.~1--8. IEEE (2007)

\bibitem{anguelov2005scape}
Anguelov, D., Srinivasan, P., Koller, D., Thrun, S., Rodgers, J., Davis, J.:
  Scape: shape completion and animation of people. In: ACM SIGGRAPH 2005
  Papers, pp. 408--416 (2005)

\bibitem{ICP}
{Besl}, P.J., {McKay}, N.D.: A method for registration of 3-d shapes. IEEE
  Transactions on Pattern Analysis and Machine Intelligence  \textbf{14}(2),
  239--256 (Feb 1992)

\bibitem{bhatnagar2020loopreg}
Bhatnagar, B.L., Sminchisescu, C., Theobalt, C., Pons-Moll, G.: Loopreg:
  Self-supervised learning of implicit surface correspondences, pose and shape
  for 3d human mesh registration. In: Advances in Neural Information Processing
  Systems ({NeurIPS}) (December 2020)

\bibitem{FAUST}
Bogo, F., Romero, J., Loper, M., Black, M.J.: {FAUST}: Dataset and evaluation
  for {3D} mesh registration. In: Proceedings IEEE Conf. on Computer Vision and
  Pattern Recognition (CVPR). IEEE, Piscataway, NJ, USA (Jun 2014)

\bibitem{chang2015shapenet}
Chang, A.X., Funkhouser, T., Guibas, L., Hanrahan, P., Huang, Q., Li, Z.,
  Savarese, S., Savva, M., Song, S., Su, H., et~al.: Shapenet: An
  information-rich 3d model repository. arXiv preprint arXiv:1512.03012  (2015)

\bibitem{keops}
Charlier, B., Feydy, J., Glaunès, J.A., Collin, F.D., Durif, G.: Kernel
  operations on the gpu, with autodiff, without memory overflows. Journal of
  Machine Learning Research  \textbf{22}(74), ~1--6 (2021),
  \url{http://jmlr.org/papers/v22/20-275.html}

\bibitem{cheng2019meshgan}
Cheng, S., Bronstein, M., Zhou, Y., Kotsia, I., Pantic, M., Zafeiriou, S.:
  Meshgan: Non-linear 3d morphable models of faces. arXiv preprint
  arXiv:1903.10384  (2019)

\bibitem{devlin-etal-2019-bert}
Devlin, J., Chang, M.W., Lee, K., Toutanova, K.: {BERT}: Pre-training of deep
  bidirectional transformers for language understanding. In: Proceedings of the
  2019 Conference of the North {A}merican Chapter of the Association for
  Computational Linguistics: Human Language Technologies, Volume 1 (Long and
  Short Papers). pp. 4171--4186. Association for Computational Linguistics,
  Minneapolis, Minnesota (Jun 2019). \doi{10.18653/v1/N19-1423},
  \url{https://www.aclweb.org/anthology/N19-1423}

\bibitem{donati2020deep}
Donati, N., Sharma, A., Ovsjanikov, M.: Deep geometric functional maps: Robust
  feature learning for shape correspondence. In: IEEE Conference on Computer
  Vision and Pattern Recognition (CVPR) (June 2020)

\bibitem{dosovitskiy2020image}
Dosovitskiy, A., Beyer, L., Kolesnikov, A., Weissenborn, D., Zhai, X.,
  Unterthiner, T., Dehghani, M., Minderer, M., Heigold, G., Gelly, S.,
  Uszkoreit, J., Houlsby, N.: An image is worth 16x16 words: Transformers for
  image recognition at scale (2020)

\bibitem{eisenberger2020smooth}
Eisenberger, M., Lahner, Z., Cremers, D.: Smooth shells: Multi-scale shape
  registration with functional maps. In: Proceedings of the IEEE/CVF Conference
  on Computer Vision and Pattern Recognition. pp. 12265--12274 (2020)

\bibitem{eisenberger2020deep}
Eisenberger, M., Toker, A., Leal-Taix{\'e}, L., Cremers, D.: Deep shells:
  Unsupervised shape correspondence with optimal transport. arXiv preprint
  arXiv:2010.15261  (2020)

\bibitem{engel2020point}
Engel, N., Belagiannis, V., Dietmayer, K.: Point transformer (2020)

\bibitem{ezuz2017deblurring}
Ezuz, D., Ben-Chen, M.: Deblurring and denoising of maps between shapes.
  Computer Graphics Forum  \textbf{36}(5),  165--174 (2017)

\bibitem{groueix20183d}
Groueix, T., Fisher, M., Kim, V.G., Russell, B.C., Aubry, M.: 3d-coded: 3d
  correspondences by deep deformation. In: Proceedings of the European
  Conference on Computer Vision (ECCV). pp. 230--246 (2018)

\bibitem{pct_guo}
Guo, M.H., Cai, J.X., Liu, Z.N., Mu, T.J., Martin, R.R., Hu, S.M.: Pct: Point
  cloud transformer. Computational Visual Media pp. 187--99 (June 2021).
  \doi{10.1007/s41095-021-0229-5}

\bibitem{hirose2020CPDbayesian}
Hirose, O.: A bayesian formulation of coherent point drift. IEEE Transactions
  on Pattern Analysis and Machine Intelligence pp.~1--1 (2020).
  \doi{10.1109/TPAMI.2020.2971687}

\bibitem{hirshberg2012coregistration}
Hirshberg, D.A., Loper, M., Rachlin, E., Black, M.J.: Coregistration:
  Simultaneous alignment and modeling of articulated 3d shape. In: European
  conference on computer vision. pp. 242--255. Springer (2012)

\bibitem{Huang_2020_CVPR}
Huang, X., Mei, G., Zhang, J.: Feature-metric registration: A fast
  semi-supervised approach for robust point cloud registration without
  correspondences. In: Proceedings of the IEEE/CVF Conference on Computer
  Vision and Pattern Recognition (CVPR) (June 2020)

\bibitem{jaegle2021perceiver}
Jaegle, A., Gimeno, F., Brock, A., Zisserman, A., Vinyals, O., Carreira, J.:
  Perceiver: General perception with iterative attention (2021)

\bibitem{jian2010robust}
Jian, B., Vemuri, B.C.: Robust point set registration using gaussian mixture
  models. IEEE transactions on pattern analysis and machine intelligence
  \textbf{33}(8),  1633--1645 (2010)

\bibitem{kim2011blended}
Kim, V.G., Lipman, Y., Funkhouser, T.: Blended intrinsic maps. ACM Transactions
  on Graphics (TOG)  \textbf{30}(4), ~79 (2011)

\bibitem{kingmaAdam}
Kingma, D.P., Ba, J.: Adam: {A} method for stochastic optimization. In: Bengio,
  Y., LeCun, Y. (eds.) 3rd International Conference on Learning
  Representations, {ICLR} 2015, San Diego, CA, USA, May 7-9, 2015, Conference
  Track Proceedings (2015), \url{http://arxiv.org/abs/1412.6980}

\bibitem{NRICP}
Li, H., Sumner, R.W., Pauly, M.: Global correspondence optimization for
  non-rigid registration of depth scans. Computer graphics forum
  \textbf{27}(5),  1421--1430 (2008)

\bibitem{litany2017deep}
Litany, O., Remez, T., Rodol{\`a}, E., Bronstein, A., Bronstein, M.: Deep
  functional maps: Structured prediction for dense shape correspondence. In:
  Proceedings of the IEEE International Conference on Computer Vision. pp.
  5659--5667 (2017)

\bibitem{loper2015smpl}
Loper, M., Mahmood, N., Romero, J., Pons-Moll, G., Black, M.J.: Smpl: A skinned
  multi-person linear model. ACM transactions on graphics (TOG)
  \textbf{34}(6),  1--16 (2015)

\bibitem{FARM}
Marin, R., Melzi, S., Rodol\`a, E., Castellani, U.: Farm: Functional automatic
  registration method for 3d human bodies. Computer Graphics Forum
  \textbf{39}(1),  160--173 (2020)

\bibitem{marin2019high}
Marin, R., Melzi, S., Rodol{\`a}, E., Castellani, U.: High-resolution
  augmentation for automatic template-based matching of human models. In: 2019
  International Conference on 3D Vision (3DV). pp. 230--239. IEEE (2019)

\bibitem{LIE2020}
Marin, R., Rakotosaona, M.J., Melzi, S., Ovsjanikov, M.: Correspondence
  learning via linearly-invariant embedding. In: Larochelle, H., Ranzato, M.,
  Hadsell, R., Balcan, M.F., Lin, H. (eds.) Advances in Neural Information
  Processing Systems. vol.~33, pp. 1608--1620. Curran Associates, Inc. (2020)

\bibitem{SHREC19}
Melzi, S., Marin, R., Rodol\`a, E., Castellani, U., Ren, J., Poulenard, A.,
  Wonka, P., Ovsjanikov, M.: {Matching Humans with Different Connectivity}. In:
  Biasotti, S., Lavoué, G., Veltkamp, R. (eds.) Eurographics Workshop on 3D
  Object Retrieval. The Eurographics Association (2019)

\bibitem{CMH}
Melzi, S., Marin, R., Musoni, P., Bardon, F., Tarini, M., Castellani, U.:
  Intrinsic/extrinsic embedding for functional remeshing of 3d shapes.
  Computers \& Graphics  \textbf{88},  1 -- 12 (2020)

\bibitem{melzi2019zoomout}
Melzi, S., Ren, J., Rodol{\`a}, E., Sharma, A., Wonka, P., Ovsjanikov, M.:
  Zoomout: Spectral upsampling for efficient shape correspondence. ACM
  Transactions on Graphics (TOG)  \textbf{38}(6), ~155 (2019)

\bibitem{myronenko2010point}
Myronenko, A., Song, X.: Point set registration: Coherent point drift. IEEE
  transactions on pattern analysis and machine intelligence  \textbf{32}(12),
  2262--2275 (2010)

\bibitem{nogneng2017informative}
Nogneng, D., Ovsjanikov, M.: Informative descriptor preservation via
  commutativity for shape matching. Computer Graphics Forum  \textbf{36}(2),
  259--267 (2017)

\bibitem{ovsjanikov2012functional}
Ovsjanikov, M., Ben-Chen, M., Solomon, J., Butscher, A., Guibas, L.: Functional
  maps: a flexible representation of maps between shapes. ACM Transactions on
  Graphics (TOG)  \textbf{31}(4),  30:1--30:11 (2012)

\bibitem{ovsjanikov2017computing}
Ovsjanikov, M., Corman, E., Bronstein, M., Rodol{\`a}, E., Ben-Chen, M.,
  Guibas, L., Chazal, F., Bronstein, A.: Computing and processing
  correspondences with functional maps. In: SIGGRAPH 2017 Courses (2017)

\bibitem{pais20203dregnet}
Pais, G.D., Ramalingam, S., Govindu, V.M., Nascimento, J.C., Chellappa, R.,
  Miraldo, P.: 3dregnet: A deep neural network for 3d point registration. In:
  Proceedings of the IEEE/CVF conference on computer vision and pattern
  recognition. pp. 7193--7203 (2020)

\bibitem{pavlakos2019expressive}
Pavlakos, G., Choutas, V., Ghorbani, N., Bolkart, T., Osman, A.A., Tzionas, D.,
  Black, M.J.: Expressive body capture: 3d hands, face, and body from a single
  image. In: Proceedings of the IEEE/CVF Conference on Computer Vision and
  Pattern Recognition. pp. 10975--10985 (2019)

\bibitem{qi2017pointnet}
Qi, C.R., Su, H., Mo, K., Guibas, L.J.: Pointnet: Deep learning on point sets
  for 3d classification and segmentation. In: Proceedings of the IEEE
  Conference on Computer Vision and Pattern Recognition. pp. 652--660 (2017)

\bibitem{radford2018improving}
Radford, A., Narasimhan, K., Salimans, T., Sutskever, I.: Improving language
  understanding by generative pre-training  (2018)

\bibitem{COMA:ECCV18}
Ranjan, A., Bolkart, T., Sanyal, S., Black, M.J.: Generating {3D} faces using
  convolutional mesh autoencoders. In: European Conference on Computer Vision
  (ECCV). pp. 725--741 (2018)

\bibitem{MapTree}
Ren, J., Melzi, S., Ovsjanikov, M., Wonka, P.: Maptree: Recovering multiple
  solutions in the space of maps. ACM Trans. Graph.  \textbf{39}(6) (Nov 2020)

\bibitem{sahilliouglu2020recent}
Sahillio{\u{g}}lu, Y.: Recent advances in shape correspondence. The Visual
  Computer  \textbf{36}(8),  1705--1721 (2020)

\bibitem{sarode2019pcrnet}
Sarode, V., Li, X., Goforth, H., Aoki, Y., Srivatsan, R.A., Lucey, S., Choset,
  H.: Pcrnet: Point cloud registration network using pointnet encoding. arXiv
  preprint arXiv:1908.07906  (2019)

\bibitem{sharp2020diffusion}
Sharp, N., Attaiki, S., Crane, K., Ovsjanikov, M.: Diffusion is all you need
  for learning on surfaces. arXiv preprint arXiv:2012.00888  (2020)

\bibitem{van2011survey}
Van~Kaick, O., Zhang, H., Hamarneh, G., Cohen-Or, D.: A survey on shape
  correspondence. Computer graphics forum  \textbf{30}(6),  1681--1707 (2011)

\bibitem{varol17_surreal}
Varol, G., Romero, J., Martin, X., Mahmood, N., Black, M.J., Laptev, I.,
  Schmid, C.: Learning from synthetic humans. In: CVPR (2017)

\bibitem{AttentionIsAllYouNeed}
Vaswani, A., Shazeer, N., Parmar, N., Uszkoreit, J., Jones, L., Gomez, A.N.,
  Kaiser, L., Polosukhin, I.: Attention is all you need. In: Proc. NIPS (2017)

\bibitem{wang2019deep}
Wang, Y., Solomon, J.M.: Deep closest point: Learning representations for point
  cloud registration. In: Proceedings of the IEEE/CVF International Conference
  on Computer Vision. pp. 3523--3532 (2019)

\bibitem{wang2019dynamic}
Wang, Y., Sun, Y., Liu, Z., Sarma, S.E., Bronstein, M.M., Solomon, J.M.:
  Dynamic graph cnn for learning on point clouds. ACM TOG  \textbf{38}(5), ~146
  (2019)

\bibitem{xu2020ghum}
Xu, H., Bazavan, E.G., Zanfir, A., Freeman, W.T., Sukthankar, R., Sminchisescu,
  C.: Ghum \& ghuml: Generative 3d human shape and articulated pose models. In:
  Proceedings of the IEEE/CVF Conference on Computer Vision and Pattern
  Recognition. pp. 6184--6193 (2020)

\bibitem{zhao2020point}
Zhao, H., Jiang, L., Jia, J., Torr, P., Koltun, V.: Point transformer (2020)

\bibitem{zuffi2015stitched}
Zuffi, S., Black, M.J.: The stitched puppet: A graphical model of 3d human
  shape and pose. In: Proceedings of the IEEE Conference on Computer Vision and
  Pattern Recognition. pp. 3537--3546 (2015)

\bibitem{smal_dataset}
Zuffi, S., Kanazawa, A., Jacobs, D.W., Black, M.J.: 3d menagerie: Modeling the
  3d shape and pose of animals. In: Proceedings of the IEEE conference on
  computer vision and pattern recognition. pp. 6365--6373 (2017)

\end{thebibliography}
\clearpage
\appendix

\section{Architecture and Implementation Details}
 
In this section, we describe in detail the proposed architecture and its implementation.

Our architecture is composed by an encoder and a decoder.

The encoder receives as input a predefined number of learnable latent probes $LP$, together with the point coordinates of the target point cloud $X_\T$. Each layer of the encoder performs an operation of cross-attention between $LP$ and $X_\T$ followed by a self-attention on $LP$. Each attention is followed by a feed-forward layer.
Before the cross-attention, input 3d-coordinates are embedded in a higher dimensional space through an MLP. We also use positional encoding on $LP$. The output of the encoder is a list of latent vectors $LS_\T$ of the same size of the input latent probes.

Specularly, the decoder receives as input the source point cloud 3d coordinates $X_\S$ and the latent vectors $LS_\T$, and is composed by layers that perform operations of cross-attention between $X_\S$ and $LS_\T$ and self-attention on $X_\S$, each followed by a feed-forward layer. $X_\S$ are also embedded in a higher dimensional space by an MLP layer, which, conversely from the encoder, shares weights among all layers. We also use positional encoding on $LS_\T$. The output of the decoder goes through a final MLP before outputting the new 3d coordinates of the input pointcloud $X_\S$ registered to the target $X_\T$.

In our implementation we use 32 latent probes of dimension 64. Point embedders are composed by four linear layers  of size $(8, 16, 32, 64)$ interleaved by ReLU activation. As standard in transformers, the feed--forward layer is made of two linear layers of size $(512, 64)$ interleaved by ReLUs.
The final MLP block is composed of five linear layers of decreasing sizes: $(48, 24, 12, 6, 3)$ also interleaved by ReLUs. We also use multi-head attention with 4 heads.
The encoder and decoder blocks, light grey in Figure 2 in the main paper, are repeated $8$ times each, weights are not shared.

When performing matching, we switch the target and source shape and we pick the version minimizing the registration's chamfer distance error.

The model was trained using an NVIDIA 2080ti. 
Keops \cite{keops} was used to improve the scalability capacity of the model.
In fact, thanks to its peculiar memory efficient implementation, it enables the processing of shapes with more than 200 thousand points (SHREC'19 dataset).

\section{Refinement Procedure}

As described in the Methods section in the main paper, the peculiar structure of our architecture allows us to refine the registration results. Here we provide additional details on this procedure.

\begin{figure}[h]  
\begin{center}
\begin{overpic}
[trim=0cm -0.9cm 0cm 0cm, clip,width=0.7\linewidth]{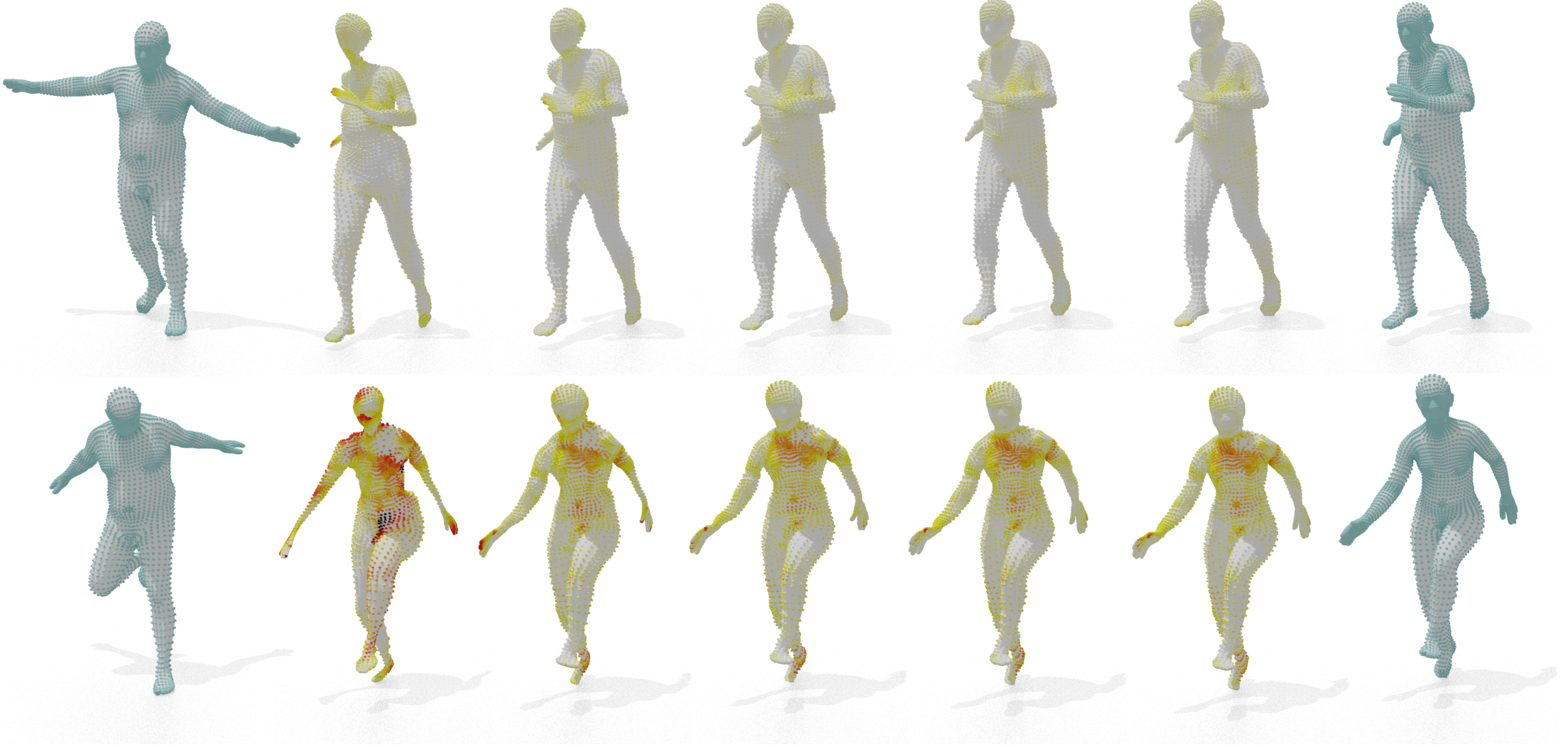}
\put(4.7,0.75){\footnotesize $X_\S$}
\put(17,0.75){\footnotesize 0 steps}
\put(32,0.75){\footnotesize 25 steps}
\put(46,0.75){\footnotesize 50 steps}
\put(60,0.75){\footnotesize 100 steps}
\put(75,0.75){\footnotesize 200 steps}
\put(92,0.75){\footnotesize $X_\T$}
\end{overpic}
\begin{minipage}{0.29\linewidth}
        \vspace{-3.7cm}
\begin{tikzpicture}
\scriptsize
\definecolor{color0}{rgb}{0.12156862745098,0.466666666666667,0.705882352941177}

\begin{axis}[
tick align=outside,
width=1.1\linewidth,
height=1.1\linewidth,
tick pos=left,
x grid style={white!69.0196078431373!black},
xlabel={refinement steps},
xmajorgrids,
xmin=0, xmax=100,
xtick style={color=black},
y grid style={white!69.0196078431373!black},
ylabel={geodesic error},
y label style={at={(0.15,0.5)}},
ymajorgrids,
ymin=0.0254135, ymax=0.0427165,
ytick style={color=black}
]
\addplot [semithick, color0, line width=2.0pt]
table {%
0 0.04193
1 0.03974
10 0.03423
20 0.03074
30 0.02933
40 0.02853
50 0.02786
60 0.02796
70 0.02722
80 0.02695
90 0.02633
100 0.0262
};
\end{axis}

\end{tikzpicture} 
\end{minipage}
\end{center} 
\caption{\label{fig:refined} On the left, a pair of qualitative examples showing the registration quality improvement using an increasing number of refinement steps of our algorithm. On the right, the mean geodesic error of our method with different number of refinement steps on FAUST1K.}
\end{figure}
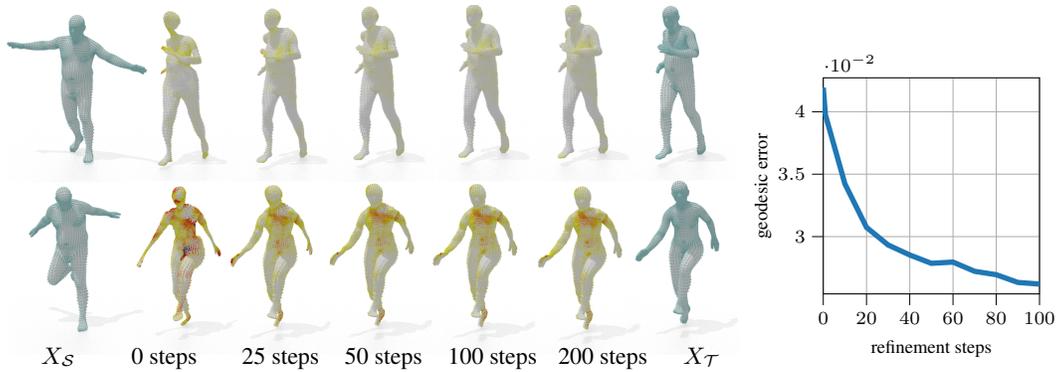

A key choice in the refinement process is the number of refinement steps. As can be seen in Figure \ref{fig:refined}, the graph suggests there is a large decline in geodesic error during the first twenty iterations, while lower marginal returns are obtained after this threshold. This is confirmed visually with the first refined registration (25 steps) having the largest decrease in error. Later iterations however contribute to more visually appealing registrations. In all experiments we perform 100 refinement steps for both the refined versions of 3DC and our method. The optimization is performed using the Adam optimizer with a learning rate of $5e-3$.


\section{SMAL Dataset}
We report qualitative results on animal shapes, confirming the ability to generalize to other types of shapes other than humans and rigid objects.

The SMAL \cite{smal_dataset} model provides the equivalent of SMPL for several animals. 
We employed this model to generate 20300 shapes of different animals.
We trained on 20000 examples and tested on the remaining 300 datapoints.

\begin{figure}[h!]
\begin{center}
\begin{overpic}
[trim=0cm 0cm 0cm 0cm, clip,width=1.0\linewidth]{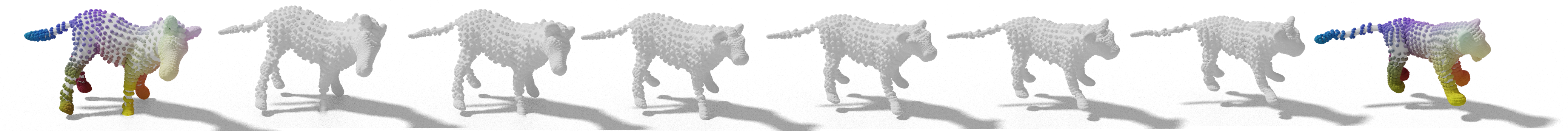}

\end{overpic}
\medskip

\begin{overpic}
[trim=0cm 0cm 0cm 0cm, clip,width=1.0\linewidth]{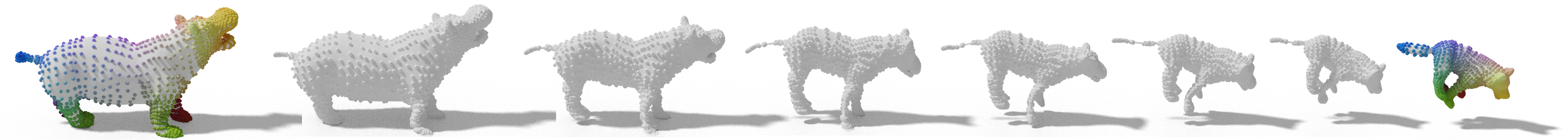}

\end{overpic}
\end{center}
\caption{\label{fig:animals} Interpolation examples on two pairs of animals belonging to different classes: horse-lion (first row), hippo-dog (second row). These results are possible only thanks to the high-quality registration provided by our method.}
\end{figure}

In Figure \ref{fig:animals}, we can observe the results we obtain with our method on this dataset.
On the extremities we have the source (left) and the target (right) shapes, colored accordingly to the predicted matching between the two. 
The shapes in the middle are gradual interpolations of the two, obtained by linearly interpolating their latent representation.
Thanks to the great flexibility of our architecture, we can generate smooth transitions and obtain high quality matching between animals of different classes in different poses.

\section{Outliers}

We have tested our model resilience to outliers using the same setup and data proposed in LinInv \cite{LIE2020}, consisting in a strong Gaussian perturbation (standard deviation of 0.03) applied to the point positions of the input point clouds.

We report in Table \ref{tab:outliers} quantitative performance for all the competitors in this setup, showing that our method is resilient to noise and manages to outperform all other methods.
We also show some qualitative results in Figure \ref{fig:s2s_reg} (last row) and in Figure \ref{fig:s2s_matching} (last two rows).

\begin{table}[h!]
\begin{center}
\caption{\label{tab:outliers} Comparison of the average geodesic error on FAUST(1k-outliers). This dataset is obtained  sampling 1k points from FAUST and perturbing each vertex with Gaussian noise with high standard deviation ($0.03$).}
\small
\begin{tabular}{lc}
\toprule
Method  & FAUST(1k-outliers)  \\ 
\toprule
3DC               & 0.2306 \\
DiffNet           & 0.3509 \\
LinInv            & 0.1738 \\
\textbf{Our}      & \textbf{0.1657} \\
\midrule 
3DC$_R$           & 0.2101 \\
\textbf{Our$_R$}  & \textbf{0.1479} \\
\bottomrule
\end{tabular}
\end{center}
\end{table}

\section{Additional Results and Visualizations}
In this section, we report further qualitative results and comparisons with other methods.

\begin{figure}[t]
\begin{center}
\begin{overpic}
[trim=0cm 0cm 0cm 0cm, clip,width=0.9\linewidth]{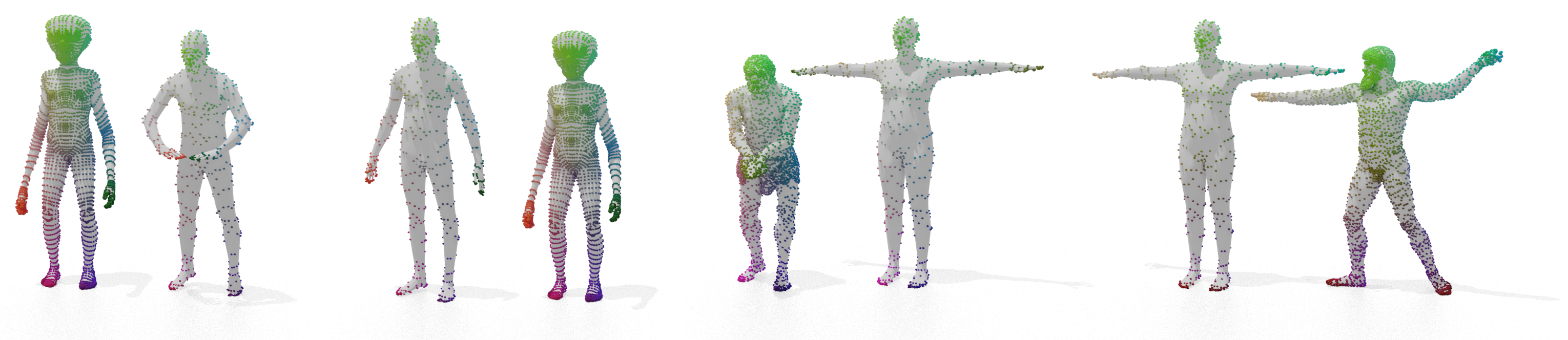}
\end{overpic}
\begin{overpic}
[trim=0cm 0cm 15cm 0cm, clip,width=0.25\linewidth]{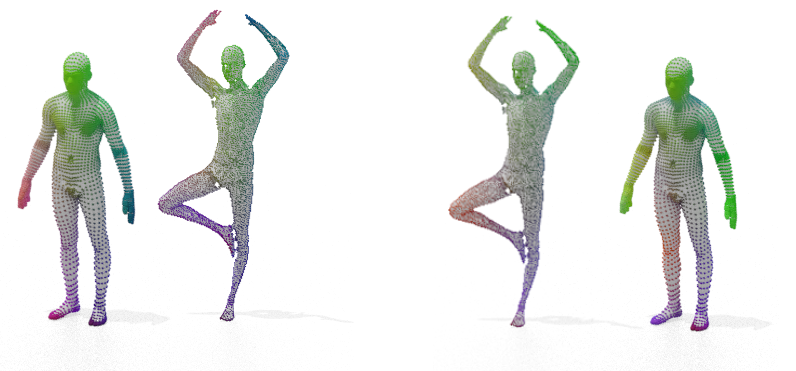}
\end{overpic}
\end{center}
\caption{\label{fig:partial} Qualitative results on shapes that present significant differences to the training examples. 
We report registrations between a FAUST shape and an alien from TOSCA, two statues from the Scan the World project, and a partial shape (second row).}
\end{figure}

\paragraph{Out of distribution samples}
In Figure \ref{fig:partial} we show results involving shapes coming from different datasets, being significantly different from the ones observed during training. In particular, we show a registration from TOSCA's alien and a FAUST shape, and vice-versa. We see that our method provides a good correspondence in both cases. Then, we also tested on two statues from the Scan the World project \cite{scantheworld}, these two present clutters and topological noises (e.g. the glued hands of the first or the beard of the second). In the second row of the same image, we report an experiment with partiality (the backside of the shape is absent). Notice that partial shapes were not seen at training time. 

\begin{figure}[h!]
\begin{center}
\begin{overpic}
[trim=0cm 0cm 0cm 0cm, clip,width=1.0\linewidth]{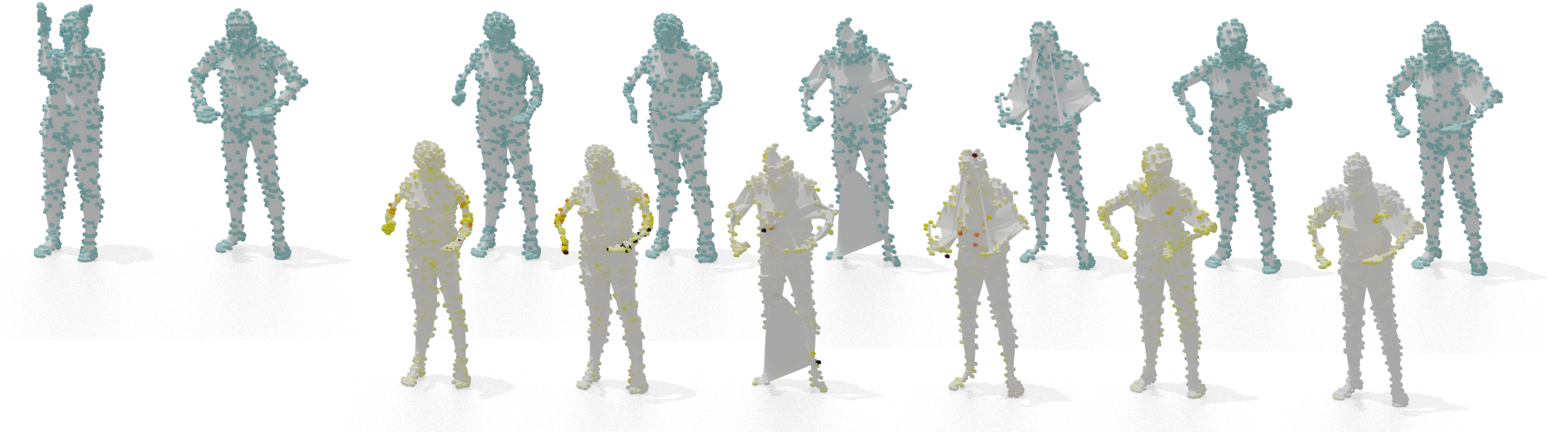}
\put(2.8,8){\footnotesize Source}
\put(14,8){\footnotesize GT} 
\put(25.5,-0.2){\footnotesize 3DC}
\put(36.5,-0.2){\footnotesize 3DC$_R$}
\put(47,-0.2){\footnotesize DiffNet}
\put(60,-0.2){\footnotesize LinInv}
\put(73,-0.2){\footnotesize Our} 
\put(86,-0.2){\footnotesize Our$_R$}
\end{overpic}\\ 
\vspace{0.25cm}
\begin{overpic}
[trim=0cm 0cm 0cm 0cm, clip,width=1.0\linewidth]{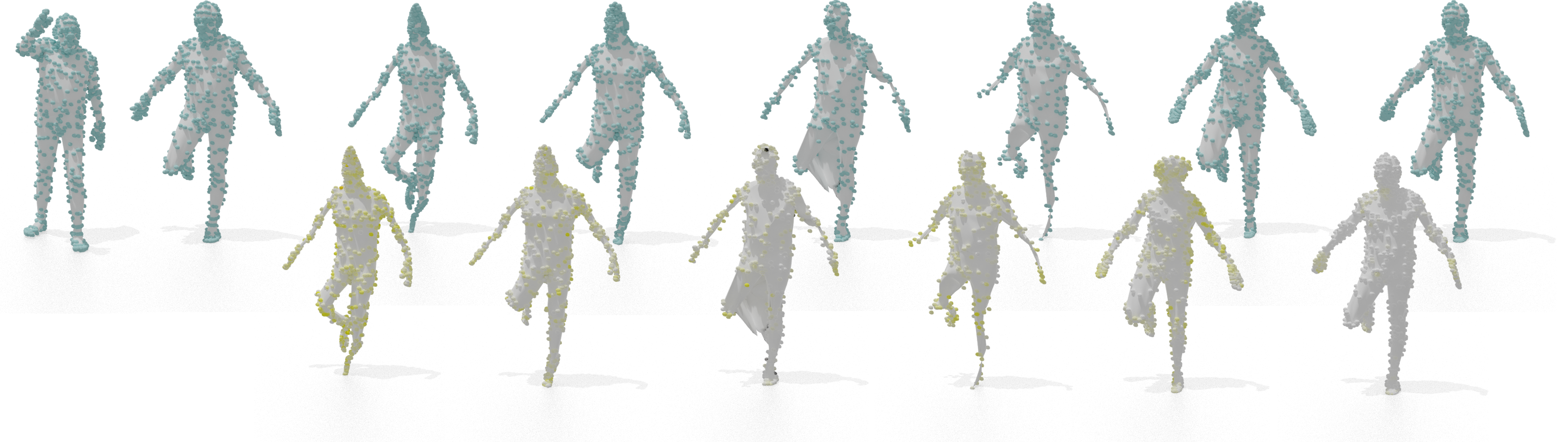}
\put(2,9.5){\footnotesize Source}
\put(11.5,9.5){\footnotesize GT} 
\put(20,0){\footnotesize 3DC}
\put(32,0){\footnotesize 3DC$_R$}
\put(46.5,0){\footnotesize DiffNet}
\put(59,0){\footnotesize LinInv}
\put(73,0){\footnotesize Our} 
\put(86,0){\footnotesize Our$_R$}
\end{overpic}\\ 
\vspace{0.25cm}
\begin{overpic}
[trim=0cm 0cm 0cm 0cm, clip,width=1.0\linewidth]{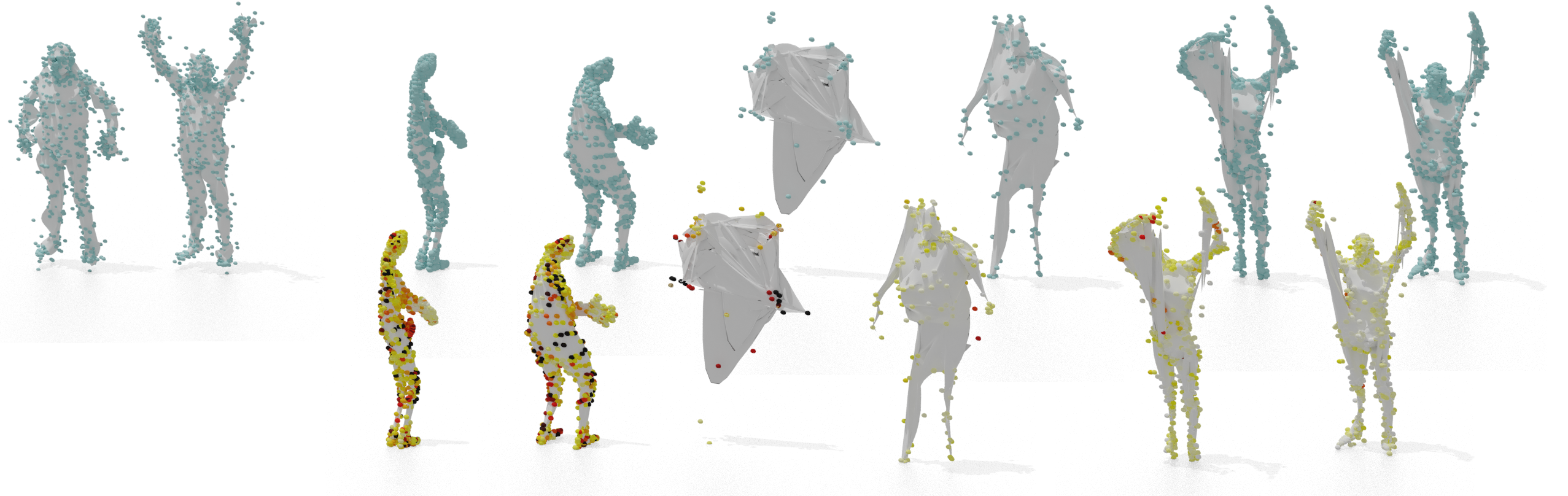}
\put(2,11){\footnotesize Source}
\put(11.5,11){\footnotesize GT} 
\put(24,0){\footnotesize 3DC}
\put(33.5,0){\footnotesize 3DC$_R$}
\put(45,0){\footnotesize DiffNet}
\put(57,0){\footnotesize LinInv}
\put(74.5,0){\footnotesize Our} 
\put(85,0){\footnotesize Our$_R$}
\end{overpic}
\end{center}
\caption{\label{fig:s2s_reg}
Qualitative comparisons of arbitrary shape registration on the FAUST (1k) \cite{FAUST} dataset \emph{(first two rows)} and the dataset of outliers from LinInv \cite{LIE2020} \emph{(last row)}. 
The registration are performed between two arbitrary shapes, however, 3DC can only register the template and not the source shape into the target shape.}
\end{figure}

\begin{figure}[t!]
\begin{center}
\begin{overpic}
[trim=0cm 0cm 0cm 0cm, clip,width=1.0\linewidth]{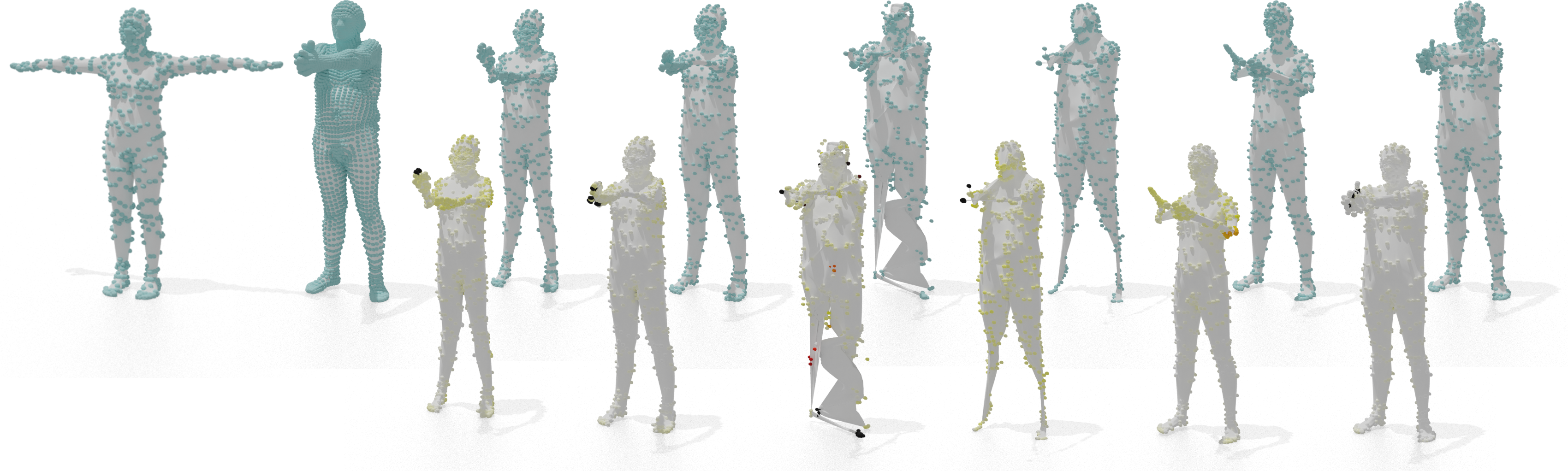}
\put(4,8){\footnotesize Source}
\put(20,8){\footnotesize GT} 
\put(27,-0.2){\footnotesize 3DC}
\put(38,-0.2){\footnotesize 3DC$_R$}
\put(49,-0.2){\footnotesize DiffNet}
\put(61,-0.2){\footnotesize LinInv}
\put(75,-0.2){\footnotesize Our} 
\put(86.5,-0.2){\footnotesize Our$_R$}
\end{overpic}
\\ 
\vspace{0.25cm}
\begin{overpic}
[trim=0cm 0cm 0cm 0cm, clip,width=1.0\linewidth]{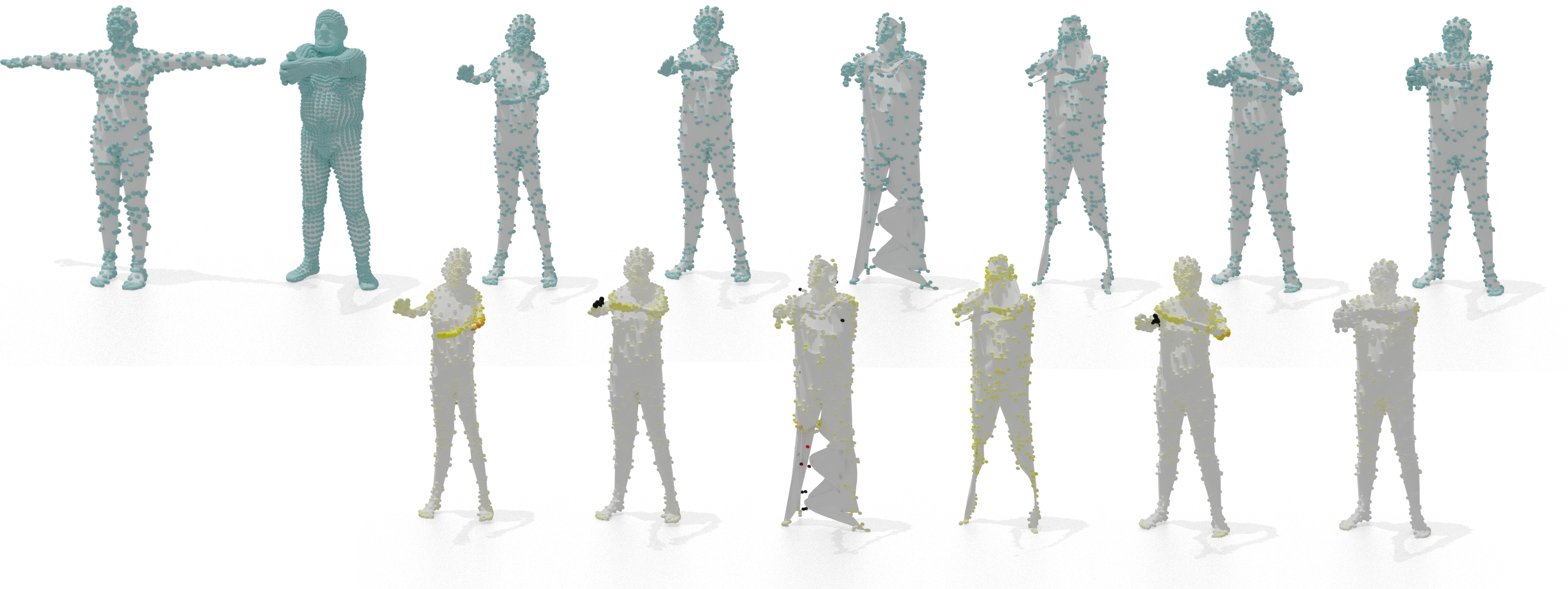}
\put(4,14){\footnotesize Source}
\put(20,14){\footnotesize GT} 
\put(27,-0.2){\footnotesize 3DC}
\put(38,-0.2){\footnotesize 3DC$_R$}
\put(49,-0.2){\footnotesize DiffNet}
\put(61,-0.2){\footnotesize LinInv}
\put(75,-0.2){\footnotesize Our} 
\put(86.5,-0.2){\footnotesize Our$_R$}
\end{overpic}
\\ 
\vspace{0.25cm}
\begin{overpic}
[trim=0cm 0cm 0cm 0cm, clip,width=1.0\linewidth]{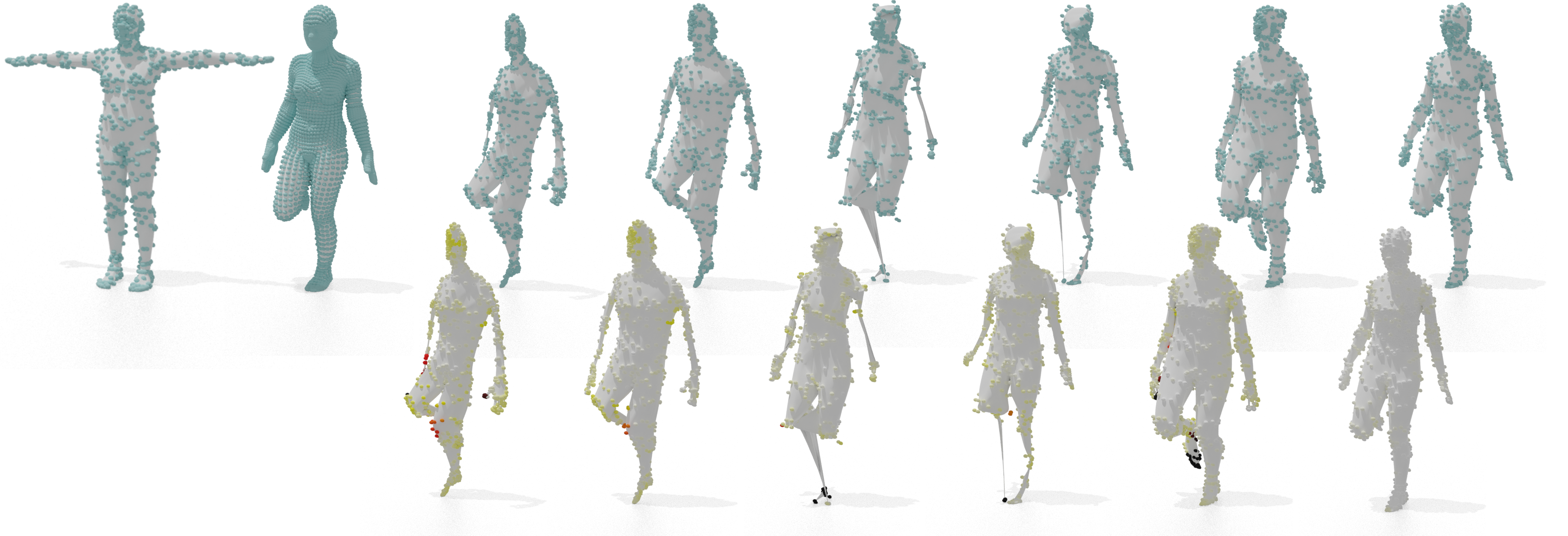}
\put(4,11){\footnotesize Source}
\put(20,11){\footnotesize GT} 
\put(27,-0.8){\footnotesize 3DC}
\put(38,-0.8){\footnotesize 3DC$_R$}
\put(49,-0.8){\footnotesize DiffNet}
\put(61,-0.8){\footnotesize LinInv}
\put(75,-0.8){\footnotesize Our} 
\put(86.5,-0.8){\footnotesize Our$_R$}
\end{overpic}
\end{center}
\caption{\label{fig:reg_s2t_full} Qualitative comparisons of template registration on FAUST.}
\end{figure}

\begin{figure}[t!]
\begin{center}
 \begin{overpic}
[trim=0cm 0cm 0cm 0cm, clip,width=1.0\linewidth]{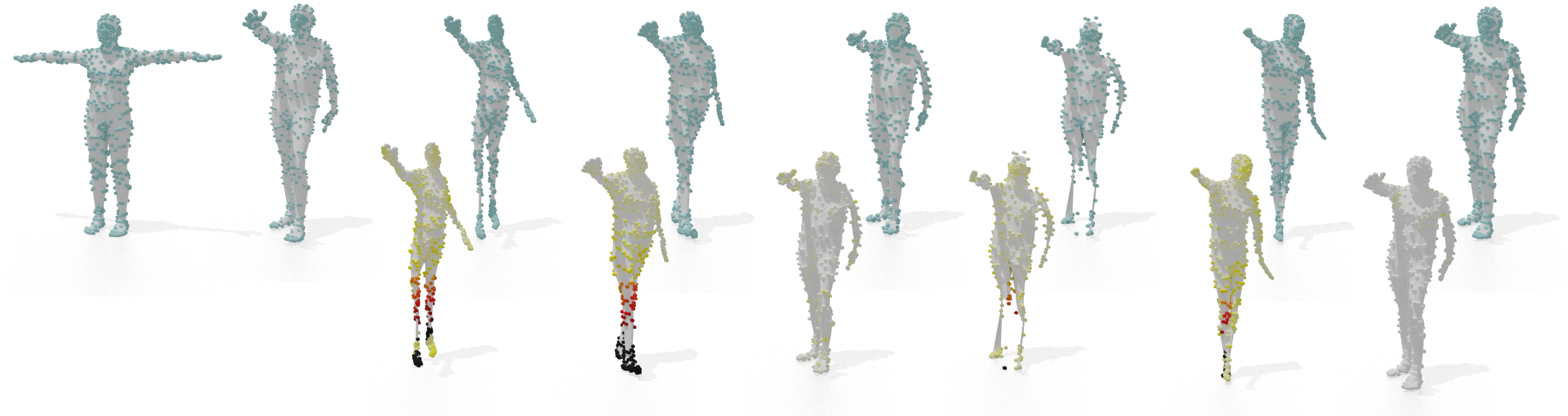}
\put(3,8){\footnotesize Source}
\put(16.5,8){\footnotesize GT} 
\put(25.5,-0.2){\footnotesize 3DC}
\put(37,-0.2){\footnotesize 3DC$_R$}
\put(48,-0.2){\footnotesize DiffNet}
\put(61,-0.2){\footnotesize LinInv}
\put(76,-0.2){\footnotesize Our} 
\put(86,-0.2){\footnotesize Our$_R$}
\end{overpic}
\\ 
\vspace{0.25cm}
 \begin{overpic}
[trim=0cm 0cm 0cm 0cm, clip,width=1.0\linewidth]{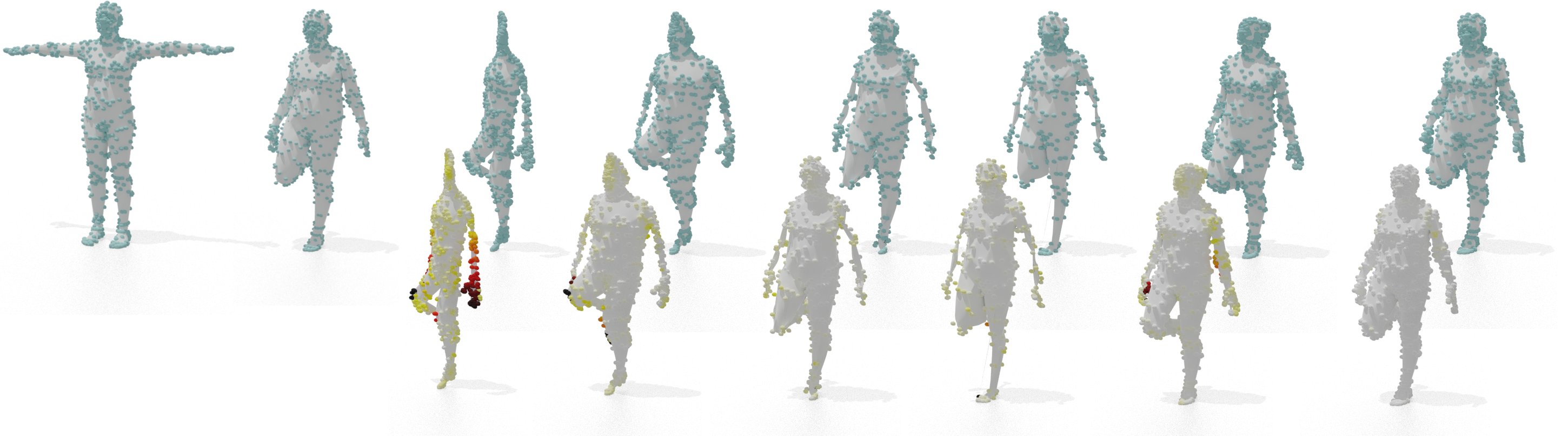}
\put(3,8){\footnotesize Source}
\put(16.5,8){\footnotesize GT} 
\put(25.5,-0.2){\footnotesize 3DC}
\put(37,-0.2){\footnotesize 3DC$_R$}
\put(48,-0.2){\footnotesize DiffNet}
\put(61,-0.2){\footnotesize LinInv}
\put(75,-0.2){\footnotesize Our} 
\put(86,-0.2){\footnotesize Our$_R$}
\end{overpic}
\\ 
\vspace{0.25cm}
 \begin{overpic}
[trim=0cm 0cm 0cm 0cm, clip,width=1.0\linewidth]{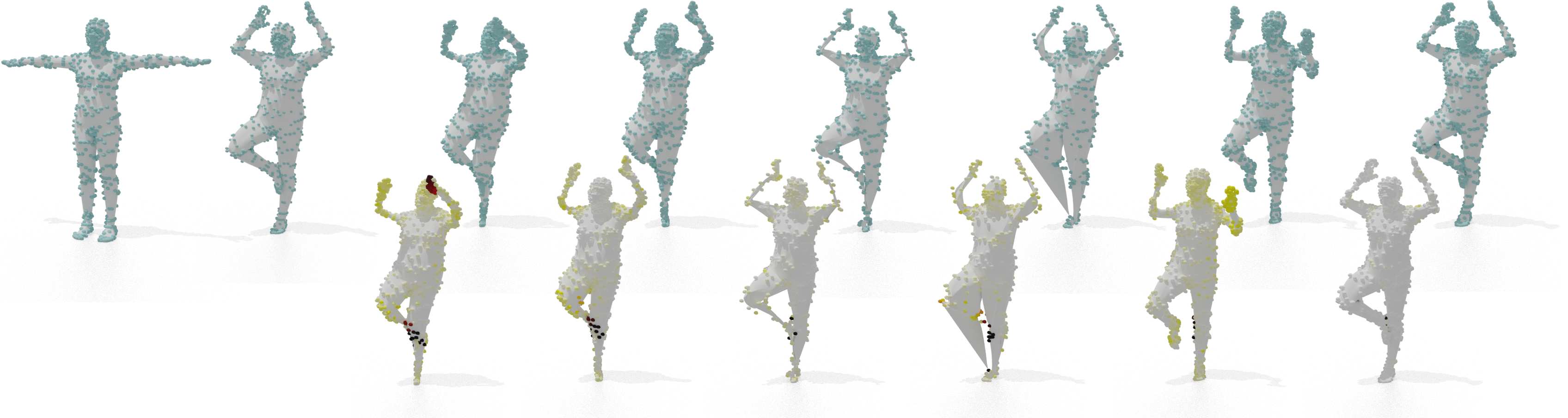}
\put(3,8){\footnotesize Source}
\put(16.5,8){\footnotesize GT} 
\put(25.5,-0.2){\footnotesize 3DC}
\put(37,-0.2){\footnotesize 3DC$_R$}
\put(48,-0.2){\footnotesize DiffNet}
\put(61,-0.2){\footnotesize LinInv}
\put(75,-0.2){\footnotesize Our} 
\put(86,-0.2){\footnotesize Our$_R$}
\end{overpic}

\end{center}
\caption{\label{fig:reg_s2t_1k} Qualitative comparisons of template registration on FAUST (1k).}
\end{figure}

\paragraph{Registration} In Figure \ref{fig:s2s_reg}, \ref{fig:reg_s2t_full} and \ref{fig:reg_s2t_1k} we show some registrations results. Notice that DiffNet and LinInv output point-to-point correspondences which naturally lie on the surface of the target shape, while registration methods have to find the correct alignment. Our registration provides a better alignment of the shapes and more accurate reconstructions of the finer details (e.g. the shape of the heads in the first row or the torso in the second one).

\begin{figure}[t!]
\begin{center}
\begin{overpic}
[trim=0cm 0cm 0cm 0cm, clip,width=1.0\linewidth]{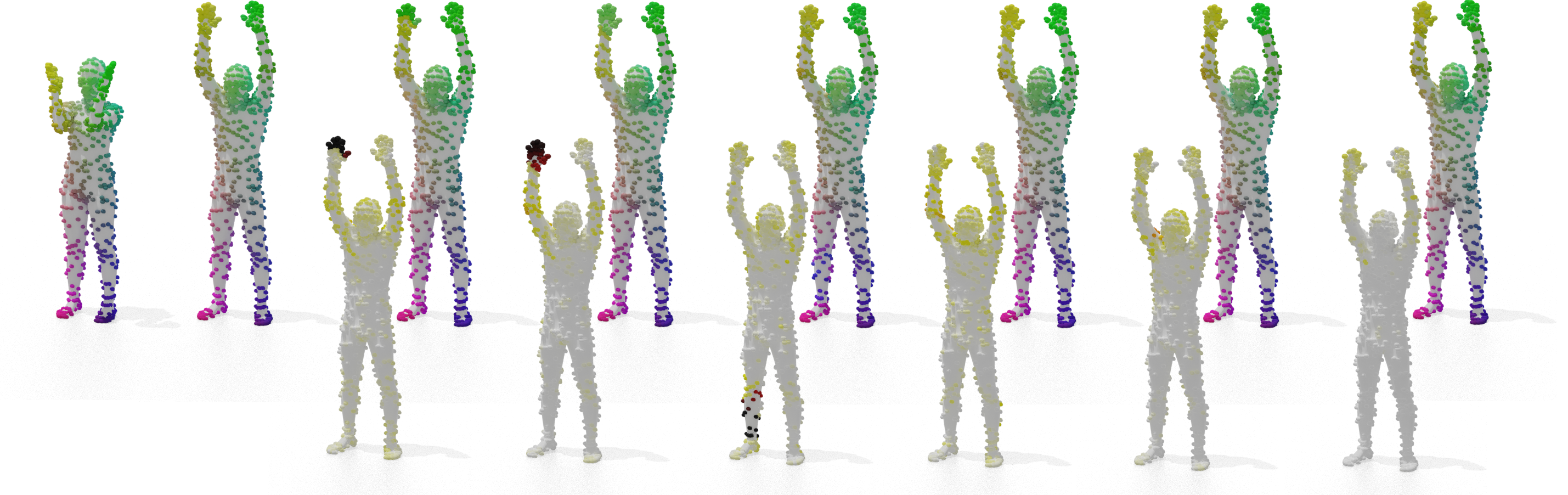}
\put(2.8,8){\footnotesize Source}
\put(13.5,8){\footnotesize GT} 
\put(21.5,0){\footnotesize 3DC}
\put(34,0){\footnotesize 3DC$_R$}
\put(46,0){\footnotesize DiffNet}
\put(59,0){\footnotesize LinInv}
\put(73,0){\footnotesize Our} 
\put(86,0){\footnotesize Our$_R$}
\end{overpic}\\ 
\vspace{0.5cm}
\begin{overpic}
[trim=0cm 0cm 0cm 0cm, clip,width=1.0\linewidth]{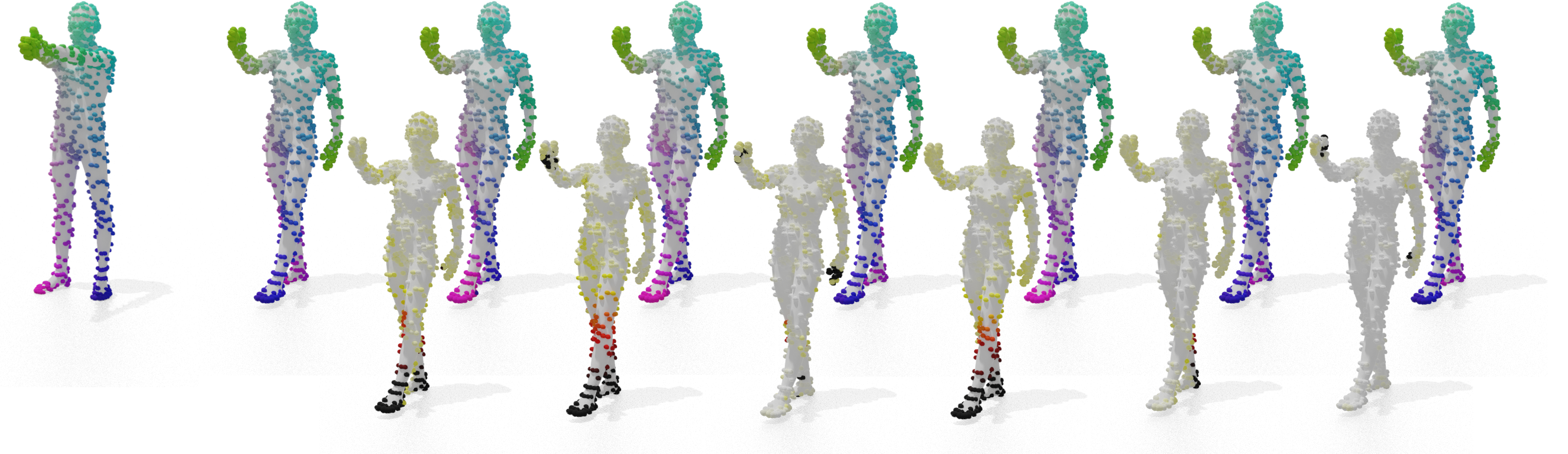}
\put(2.5,7){\footnotesize Source}
\put(17,7){\footnotesize GT} 
\put(23,0){\footnotesize 3DC}
\put(34.5,0){\footnotesize 3DC$_R$}
\put(47,0){\footnotesize DiffNet}
\put(59,0){\footnotesize LinInv}
\put(73,0){\footnotesize Our} 
\put(86,0){\footnotesize Our$_R$}
\end{overpic}\\ 
\vspace{0.5cm}
\begin{overpic}
[trim=0cm 0cm 0cm 0cm, clip,width=1.0\linewidth]{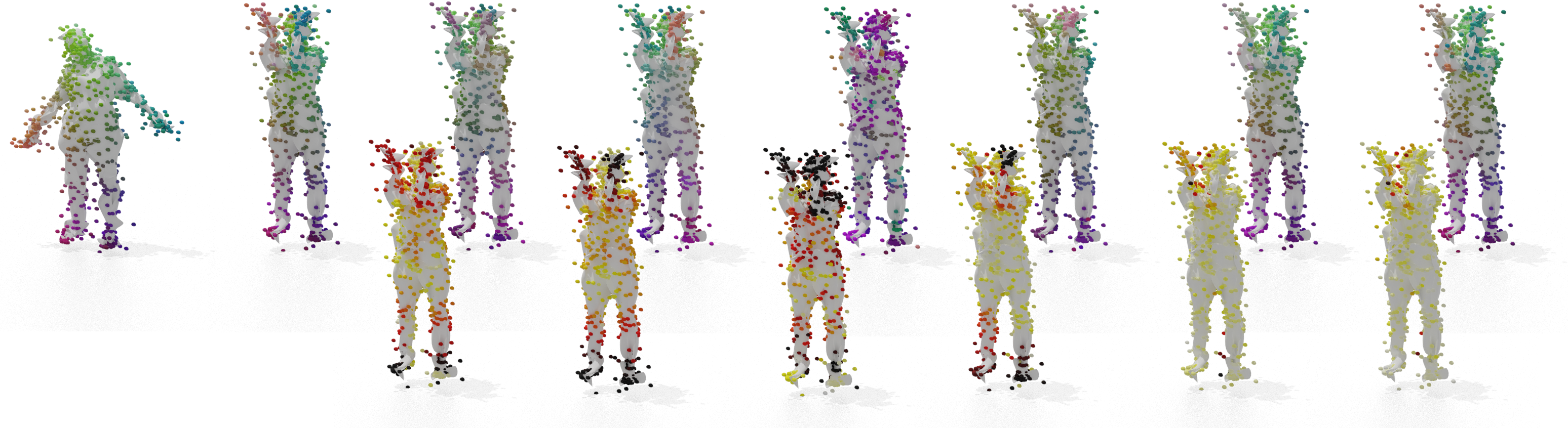}
\put(2.8,8){\footnotesize Source}
\put(17,8){\footnotesize GT} 
\put(25,0){\footnotesize 3DC}
\put(37,0){\footnotesize 3DC$_R$}
\put(50,0){\footnotesize DiffNet}
\put(61,0){\footnotesize LinInv}
\put(76,0){\footnotesize Our} 
\put(88,0){\footnotesize Our$_R$}
\end{overpic}\\ 
\vspace{0.5cm}
\begin{overpic}
[trim=0cm 0cm 0cm 0cm, clip,width=1.0\linewidth]{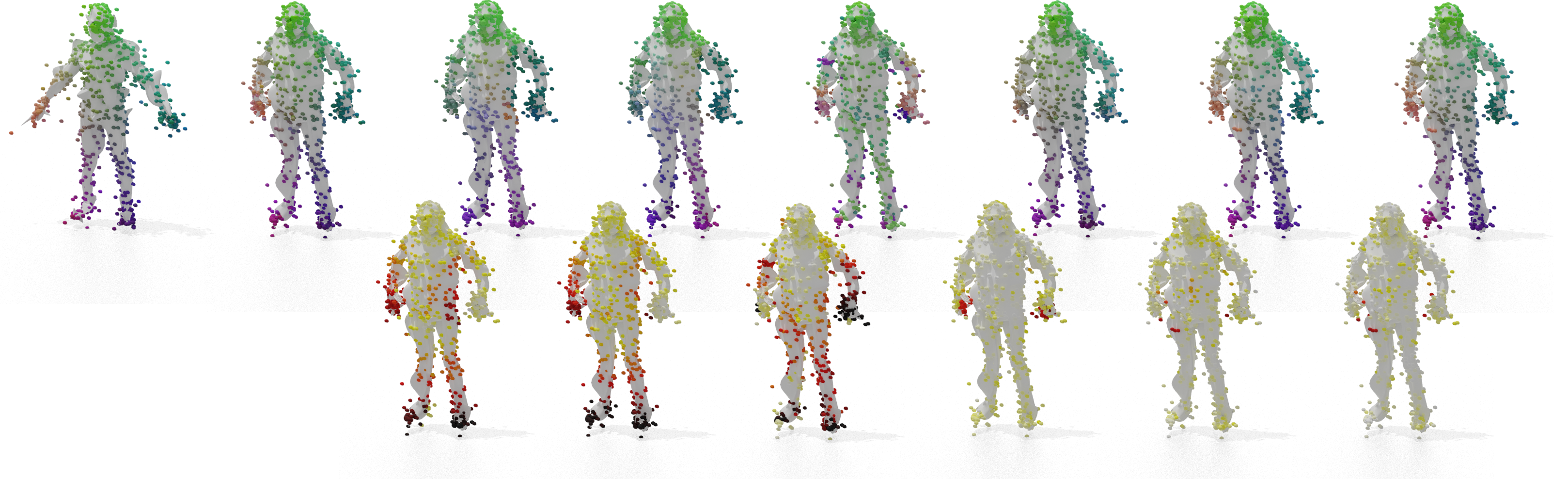}
\put(3,12){\footnotesize Source}
\put(17.5,12){\footnotesize GT} 
\put(25,0){\footnotesize 3DC}
\put(37,0){\footnotesize 3DC$_R$}
\put(50,0){\footnotesize DiffNet}
\put(61,0){\footnotesize LinInv}
\put(76,0){\footnotesize Our} 
\put(88,0){\footnotesize Our$_R$}
\end{overpic} 
\end{center}
\caption{\label{fig:s2s_matching}
Qualitative comparisons of shape matching on the FAUST (1k) \cite{FAUST} dataset \emph{(first two rows)} and the dataset of outliers from LinInv \cite{LIE2020} \emph{(last two rows)}.}
\end{figure}

\paragraph{Matching} In Figure \ref{fig:s2s_matching} we report some shape-to-shape matching on FAUST 1K (in the first two rows) and on the outliers dataset proposed in \cite{LIE2020} (in the last two rows). For each pair, we visualize the matching as color transfer, showing from left to right, the ground-truth, the outcome of 3DC without and with refinement, DiffusionNet, LinInv, and our method without and with refinement. For our method, we observe a general resilience to sparsity and noise.

\begin{figure}[t!]
\begin{center}
\begin{overpic}
[trim=0cm 0cm 0cm 0cm, clip,width=1.0\linewidth]{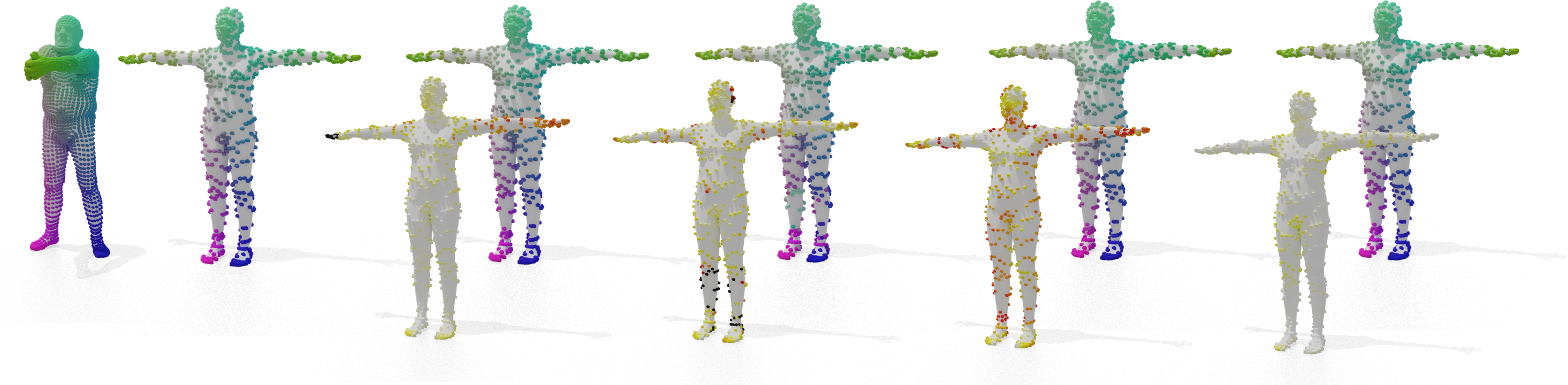}
\put(3,0.75){\footnotesize Source}
\put(13.5,0.75){\footnotesize GT}
\put(32,0.75){\footnotesize 3DC$_R$}
\put(51,0.75){\footnotesize DiffNet}
\put(69,0.75){\footnotesize LinInv}
\put(87,0.75){\footnotesize Our$_R$}
\end{overpic}
\end{center}
\caption{\label{fig:matching_s2t} A qualitative comparison of template registration on FAUST \cite{FAUST}. 
In particular, \emph{3DC$_R$} exhibits an error due to topological changes between $\S$ and $\T$, \emph{LinInv} shows small but widespread errors.}
\end{figure}

\begin{figure}[t!]
\begin{center}
\begin{overpic}
[trim=0cm 0cm 0cm 0cm,clip,width=1.0\linewidth]{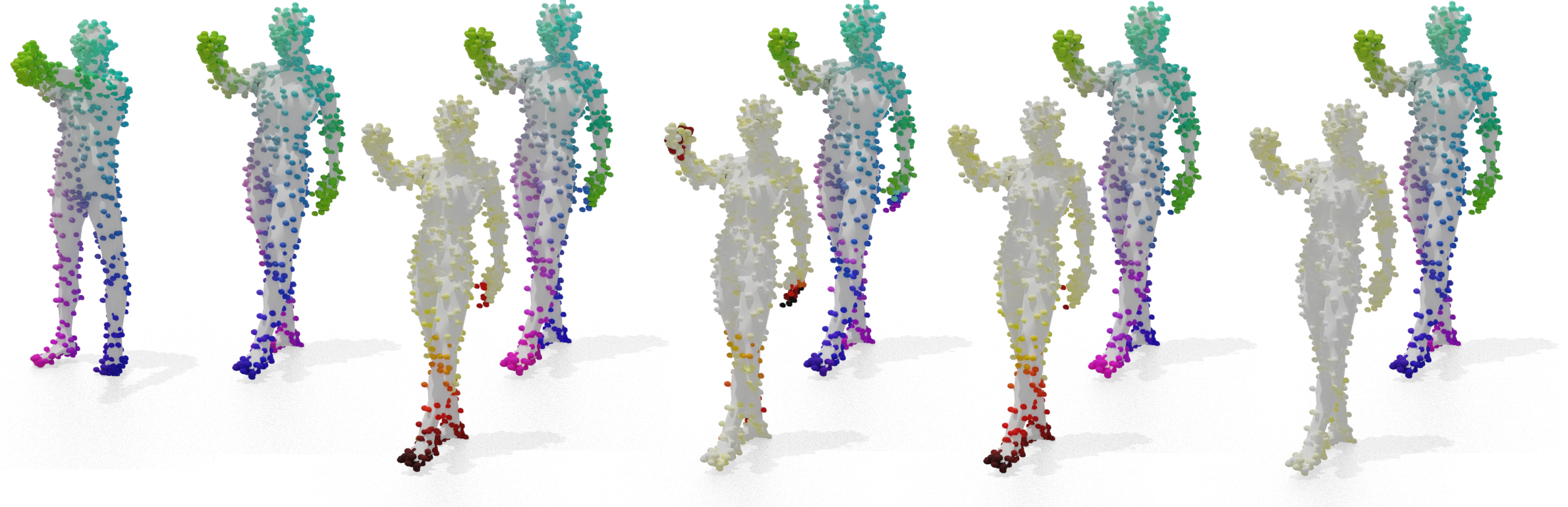}
\put(2.7,0.75){\footnotesize Source}
\put(16,0.75){\footnotesize GT}
\put(31,0.75){\footnotesize 3DC}
\put(51,0.75){\footnotesize DiffNet}
\put(69,0.75){\footnotesize LinInv}
\put(88,0.75){\footnotesize Our}
\end{overpic}
\end{center}
\caption{\label{fig:faustP} A qualitative comparison of shape matching on the perturbed FAUST \cite{FAUST} dataset. 
}
\end{figure}

In Figure \ref{fig:matching_s2t} we collect further color transferring qualitative results, with also the matching error reported as hot map on the point clouds. In Figure \ref{fig:faustP}, we visualize some results in the same setup but with additional noise on the point clouds.

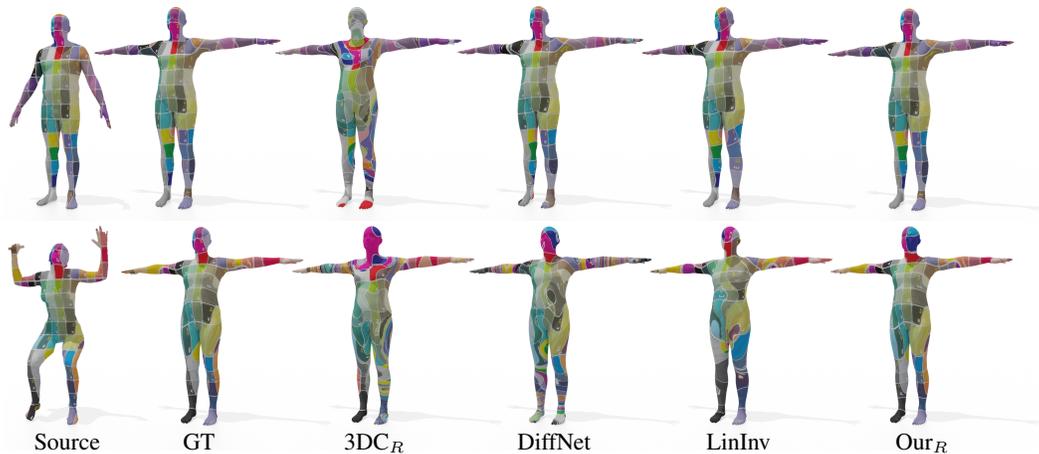
\begin{figure}[t!]
\begin{center}
\begin{overpic}
[trim=0cm 0cm 0cm 0cm,clip,width=1.0\linewidth]{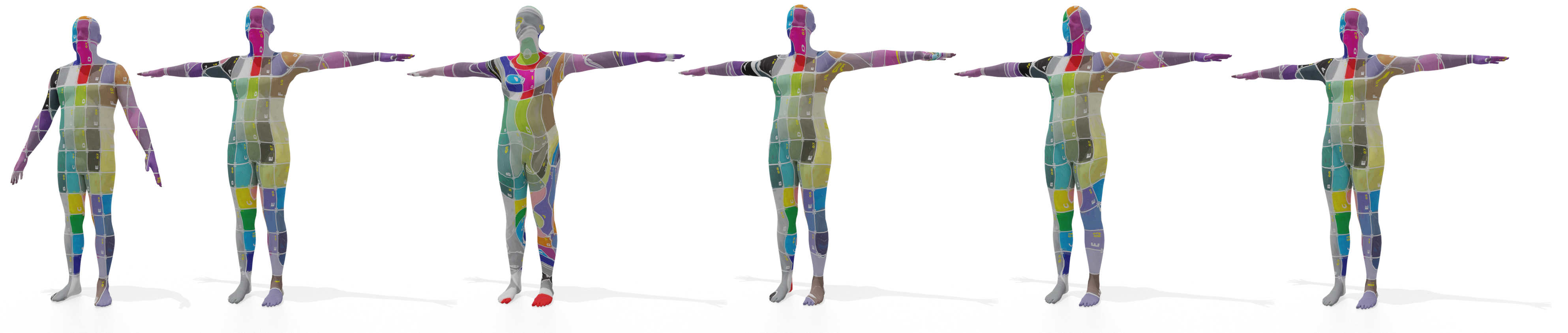}
\end{overpic}
\begin{overpic}
[trim=0cm 0cm 0cm 0cm,clip,width=1.0\linewidth]{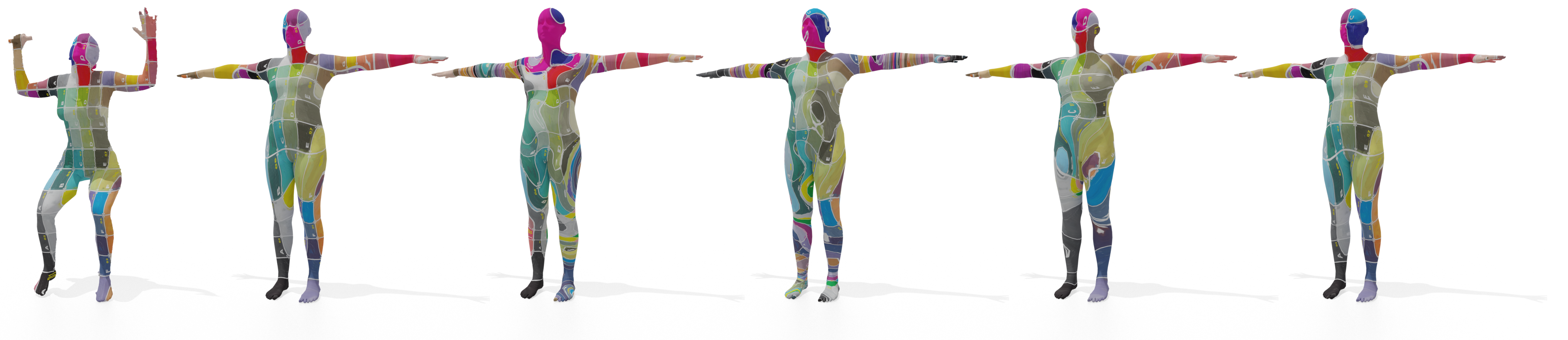}
\put(3,0){\footnotesize Source}
\put(17.2,0){\footnotesize GT}
\put(32.5,0){\footnotesize 3DC$_R$}
\put(49,0){\footnotesize DiffNet}
\put(67,0){\footnotesize LinInv}
\put(85,0){\footnotesize Our$_R$}
\end{overpic}
\end{center}
\caption{\label{fig:texture} Two qualitative comparisons of texture transfer on the SHREC19 \cite{SHREC19} dataset. 
From left to right, the source shape $\S$, the ground truth transfer to the target geometry $\T$, the results of the competitors and our result. In the top row we report an \emph{easy} example of texture transfer, where almost all competitors perform reasonably well. In the bottom row,  we report an \emph{hard} example of texture transfer caused by the large variation in the pose, the missing parts (fingers) and the presence of extra object (in the hand of the source shape).} 
\end{figure}

Finally, in Figure \ref{fig:texture} we report a texture transfer performed between two shapes from SHREC19 dataset \cite{SHREC19}.

\end{document}